\pdfoutput=1
\documentclass{article}
\usepackage{preprint,times}

\usepackage{booktabs}
\usepackage{array}
\usepackage[table]{xcolor}
\usepackage{caption}
\usepackage{amsmath}
\usepackage{amssymb}
\usepackage{graphicx}
\usepackage{subcaption}
\usepackage[pagebackref,breaklinks,linkcolor=cornellred,urlcolor=black,colorlinks=true,citecolor=teal]{hyperref}
\usepackage{cleveref}
\usepackage{algorithm}
\usepackage{algpseudocode}
\usepackage{float}
\usepackage{multirow}

\definecolor{cornellred}{rgb}{0.7, 0.11, 0.11}

\definecolor{pairbg}{HTML}{E8ECF4}
\definecolor{modelbg}{HTML}{F5F6F8}
\definecolor{oursbg}{HTML}{FDF4E3}
\definecolor{gainpos}{HTML}{1A7F37}
\definecolor{gainneg}{HTML}{B42318}
 
\newcommand{\best}[1]{\bfseries #1}
\newcommand{\up}[1]{{\color{gainpos}\scriptsize$\uparrow$#1}}
\newcommand{\dn}[1]{{\color{gainneg}\scriptsize$\downarrow$#1}}
\newcommand{\refrow}{{\scriptsize --}}

\newcommand{\oursrow}{\rowcolor{oursbg}}
\newcommand{\stdv}[1]{{\scriptsize\,$\pm$#1}}
 
\newcommand{\NC}{8}
\newcommand{\bandstrut}{\rule[-0.9ex]{0pt}{3.2ex}}
\newcommand{\pairband}[1]{%
  \rowcolor{pairbg}\multicolumn{\NC}{@{}l@{}}{\bandstrut\textbf{#1}}\\}
\newcommand{\modelband}[1]{%
  \rowcolor{modelbg}\multicolumn{\NC}{@{}l@{}}{\bandstrut\quad\textit{#1}}\\}

\newcommand{\method}{GRAFT}
\newcommand{\fullmethod}{\textbf{G}ated \textbf{R}eplacement of \textbf{A}nswer-\textbf{F}ailed groups with peer \textbf{T}rajectories}

\usepackage{amsmath,amsfonts,bm}

\def\eqref#1{equation~\ref{#1}}

\def\1{\bm{1}}

\DeclareMathAlphabet{\mathsfit}{\encodingdefault}{\sfdefault}{m}{sl}
\SetMathAlphabet{\mathsfit}{bold}{\encodingdefault}{\sfdefault}{bx}{n}

\usepackage{url}

\title{Learning Beyond What You Sample: \\Off-Policy-Aware Cross-Model \\Trajectory Exchange for RLVR}

\author{Doohyuk Jang$^{1}$, Yoonsik Park$^{1}$, Gyouk Chu$^{1}$, Sihwan Park$^{1}$, Eunho Yang$^{1,2\dagger}$ \\
$^{1}$KAIST \quad $^{2}$AITRICS
}

\begin{document}

\maketitle
\begin{NoHyper}\renewcommand{\thefootnote}{}\footnotetext{$^{\dagger}$Corresponding author: \texttt{eunhoy@kaist.ac.kr}}\end{NoHyper}

\begin{abstract}
Reinforcement Learning with Verifiable Rewards (RLVR) methods such as GRPO rely on successful self-generated trajectories, but finite rollout budgets can produce all-fail groups with no reward-based policy-gradient signal. While additional rollouts improve the chance of success at higher cost, successful trajectories missing from one model's rollouts may already have been discovered by another. Indeed, we observe that heterogeneous models often succeed on complementary prompts, creating opportunities for mutual learning without a designated stronger teacher. To exploit this complementarity, we propose \textbf{\method{}} (\fullmethod{}), an off-policy-aware framework that replaces all-fail groups with informative peer groups. \method{} transfers both successful and unsuccessful peer responses with peer-computed advantages, while controlling cross-model mismatch through sequence-level compatibility weighting and token-level importance ratio clipping. Across three heterogeneous model pairs and five mathematical reasoning benchmarks, \method{} consistently improves both models over GRPO with the same per-model rollout budget, gaining 2.1 points on average and up to 4.5 points in model-level average performance. Stored peer trajectories preserve most of the gains, improving over GRPO by 1.8 points on average without simultaneous co-training.
\end{abstract}

\section{Introduction}
\label{sec:intro}

Reinforcement Learning with Verifiable Rewards (RLVR) has become a key post-training paradigm for improving the reasoning ability of Large Language Models (LLMs), with substantial gains on many reasoning tasks~\citep{shao2024deepseekmath,lambert2025tulu,r1}.
Group Relative Policy Optimization (GRPO)~\citep{shao2024deepseekmath} and its variants~\citep{dapo,liu2025understanding,gspo,kim2026discounted} optimize the policy using relative rewards among sampled responses, raising the likelihood of high-reward trajectories and suppressing low-reward ones.
In this on-policy setting, learning reinforces successful reasoning strategies from self-generated trajectories, without fine-grained supervision~\citep{r1,wen2026reinforcement}.

The effectiveness of this learning process depends on whether the model discovers a successful trajectory within its rollout budget~\citep{liu-etal-2026-evocot,NEURIPS2025_537d5aa7,dong-etal-2026-rl}.
This limitation arises when every sampled response fails, leaving all group-relative advantages at zero.
Such groups can be dropped, their prompts replaced, or their rewards reshaped to recover a non-zero learning signal~\citep{dapo,le2026no,he2026advantage}, while increasing the group size improves the odds of sampling a correct trajectory at higher rollout cost.
These approaches, however, remain dependent on the learner's own exploration.

\begin{figure}[t]
  \centering
  \begin{subfigure}[t]{0.66\textwidth}
    \centering\includegraphics[width=\linewidth]{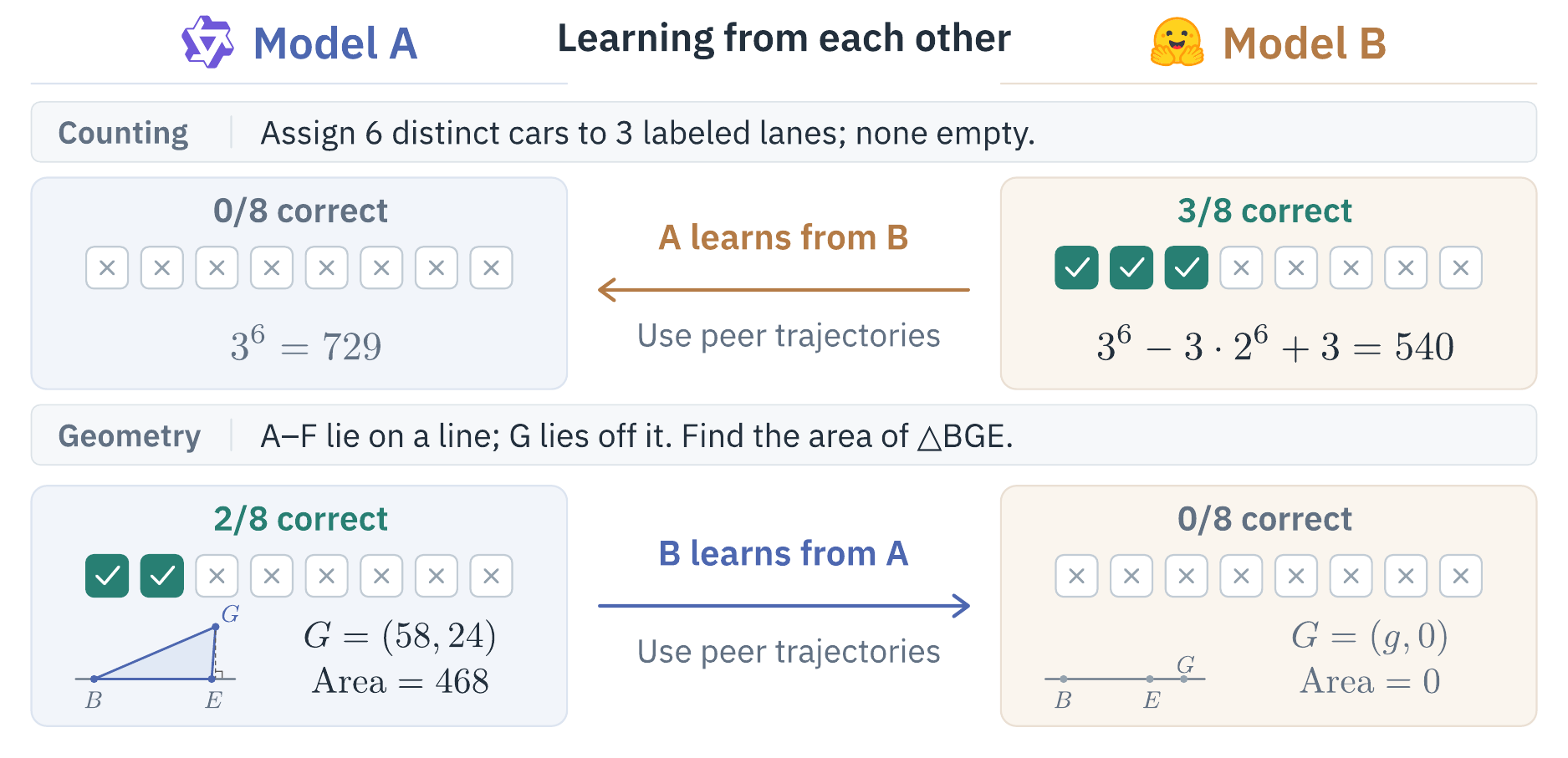}
    \label{fig:concept-fig}
  \end{subfigure}\hfill
  \begin{subfigure}[t]{0.33\textwidth}
    \centering\includegraphics[width=\linewidth]{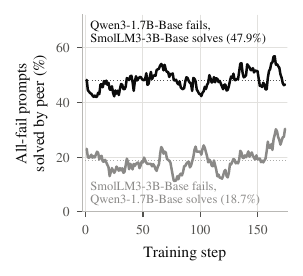}
    \label{fig:rescue-frac}
  \end{subfigure}
  \caption{\textbf{Complementary successes persist throughout training.} \textbf{(Left)} Illustrative examples where one model solves a prompt its peer fails entirely, so each could learn from the other's trajectories. \textbf{(Right)} Fraction of all-fail prompts solved by the peer, with means shown as dotted lines.}
  \label{fig:concept}
\end{figure}

The growing diversity of open-source LLMs creates an opportunity to move beyond learning solely from self-generated trajectories: a successful response missing from one model's rollouts may already be present in another's.
Yet standard single-model RLVR trains each policy in isolation, leaving these successes unused.
We observe this complementarity in independent GRPO runs of SmolLM3-3B-Base~\citep{bakouch2025smollm3} and Qwen3-1.7B-Base~\citep{qwen3technicalreport}, two models with distinct pretraining histories (\Cref{fig:concept}).
SmolLM3-3B-Base solves 47.9\% of the prompts on which Qwen3-1.7B-Base fails across all eight rollouts, while Qwen3-1.7B-Base solves 18.7\% of SmolLM3-3B-Base's all-fail prompts.
These complementary successes suggest that both models could benefit from exchanging verified solutions, without requiring a designated stronger teacher.

Turning complementary successes into learning gains requires deciding \emph{which} prompts warrant peer supervision and \emph{how} to learn from peer responses. 
These choices interact: all-fail prompts lack a reward-based policy-gradient signal, but successful peer responses may still be poorly matched to the receiver. Conversely, transferring peer responses on prompts with an informative self-generated group changes an update that already has an on-policy learning signal.
HACPO~\citep{zhang2026heterogeneous} accounts for cross-model mismatch but shares peer rollouts beyond receiver-failure prompts, while SGT~\citep{liu2026experience} targets receiver failures through a fixed-weight supervised loss without compatibility-based weighting. Applying either update rule to our selected prompts still underperforms \method{} (\Cref{sec:ablation}). This motivates pairing complementary prompt selection with compatibility-aware learning from full peer groups.

We propose \textbf{\method{}} (\fullmethod{}), an off-policy-aware framework for cross-model trajectory sharing in RLVR that addresses both decisions.
For \emph{which}, \method{} selects prompts on which the receiver fails entirely and the peer produces both successful and unsuccessful responses, while balancing exchange volume across directions.
For \emph{how}, it replaces the selected receiver groups with the corresponding peer groups and retains their source-computed advantages, preserving within-peer reward contrast without pooling rewards across models.
It controls peer influence through bounded sequence-level compatibility gating and token-level importance ratio clipping, and keeps the receiver's own data primary by processing peer-containing minibatches after its on-policy minibatches.

Across three heterogeneous pairs of open-source base models and five mathematical reasoning benchmarks, \method{} improves both models in every pair over GRPO ($n{=}8$), by 2.1 points on average and up to 4.5 points in model-level average score.
In two of the three pairs, both models match or exceed GRPO trained with four times the rollout budget ($n{=}32$).
\method{} also outperforms HACPO and SGT by 4.0 and 1.5 points on average, respectively.
The gains largely persist when reusing peer trajectories from completed independent GRPO runs ($+1.8$ points on average), without simultaneous co-training or additional peer rollouts.

\section{Related Work}
\label{sec:rel}

\paragraph{Exploration limitations in RLVR.}
GRPO and related RLVR methods learn only from reward variation within self-generated rollout groups~\citep{shao2024deepseekmath,dapo}.
Dynamic sampling discards zero-variance groups and resamples~\citep{dapo}, while larger groups improve success coverage; both cost extra rollouts without guaranteeing success.
Entropy-guided advantage shaping recovers a signal without additional rollouts~\citep{le2026no}, but on an all-incorrect group it can only suppress the sampled failures, not supply a correct response.
Even at scale, RLVR improves sampling efficiency without expanding the base model's solvable prompt set~\citep{NEURIPS2025_537d5aa7}, and declining entropy further limits exploration~\citep{cui2025entropy}.
Hints, partial solutions, and expert guidance ease exploration but require an external solution source or a stronger model~\citep{li2026questa,huang2026blending,jiang2026selective}.
We instead use trajectories from heterogeneous peers when the learner's own rollouts all fail.

\paragraph{Off-policy guidance and mismatch control.}
External demonstrations, teacher solutions, and historical trajectories augment RLVR rollouts but introduce policy mismatch~\citep{NEURIPS2025_a9d5c33e,dong-etal-2026-rl,mao2026rlvr}.
Prior work addresses rollout--training mismatch through importance weighting and truncation~\citep{yao2025on,ling2025everystep}, and studies sequence-level optimization and off-policy correction~\citep{gspo,chen2025minimax}.
We consider distinct peer models with potentially different tokenizers.
\method{} separates within-receiver policy change from cross-model mismatch through token-level importance ratio clipping and sequence-level compatibility filtering with bounded weighting.
The compatibility score is an empirical proxy from average token log-likelihoods, not an exact cross-tokenizer importance ratio.

\paragraph{Cross-model learning and trajectory sharing.}
Recent work enables multiple models to learn from one another during RL.
HACPO~\citep{zhang2026heterogeneous} exchanges peer rollouts with off-policy correction, while Mutual RL~\citep{liu2026experience} introduces SGT to transfer verified peer successes on prompts where the receiver fails.
F-TIS~\citep{blagoev2026ftis} studies collaborative GRPO among models from the same family with a shared vocabulary, using truncated importance sampling and off-policy filtering.
Unlike teacher-guided distillation~\citep{agarwal2024policy}, these approaches motivate learning across peer models without relying exclusively on a designated stronger teacher.
We build on this direction by jointly addressing where peer trajectories provide missing supervision and how their influence should be controlled under cross-model mismatch.

\section{Preliminaries}
\label{sec:prelim}

\paragraph{Group Relative Policy Optimization.}
Given a prompt \(q\) sampled from a prompt set \(\mathcal{D}\), GRPO~\citep{shao2024deepseekmath} samples a group of \(n\) responses
\(\mathcal{G}(q)=(o_1,\dots,o_n)\)
from an old policy \(\pi_{\theta_{\mathrm{old}}}\) and assigns each response a verifiable reward
\(r_i=r(q,o_i)\in\{0,1\}\).
In this section we identify each response with its token sequence and write \(o_i=(o_{i,1},\dots,o_{i,|o_i|})\); \Cref{sec:method} makes tokenizers explicit. Let \(\mathcal{R}(q)=(r_1,\dots,r_n)\) denote the corresponding rewards.
The group-relative advantage of response \(o_i\) is computed as
\begin{equation}
\hat{a}_i
=
\frac{r_i-\operatorname{mean}(\mathcal{R}(q))}
{\operatorname{std}(\mathcal{R}(q)) + \epsilon_{\mathrm{adv}}}.
\end{equation}
When \(\mathrm{std}(\mathcal{R}(q))=0\), \textit{i.e.}, the group is entirely correct or entirely incorrect, all advantages become zero, so the group contributes no policy-gradient signal.
Following DAPO~\citep{dapo}, a GRPO variant, we use token-level loss aggregation with asymmetric clipping:
\begin{equation}
\resizebox{0.92\linewidth}{!}{$\displaystyle
\mathcal{J}_{\mathrm{GRPO}}(\theta)=
\mathbb{E}_{
\mathcal{B}
}
\left[
\frac{1}{\sum_{j\in\mathcal{B}} |o_j|}
\sum_{j\in\mathcal{B}}
\sum_{t=1}^{|o_j|} 
\min\left( \rho_{j,t}(\theta)\hat{a}_j,\, 
\operatorname{clip}\!\left( 
\rho_{j,t}(\theta),1-\varepsilon_{\mathrm{low}},1+\varepsilon_{\mathrm{high}} 
\right)\hat{a}_j 
\right) 
\right],
$}
\label{eq:grpo-obj}
\end{equation}
where \(\mathcal{B}\) denotes a batch of responses sampled from \(\pi_{\theta_{\mathrm{old}}}\), with their corresponding prompts drawn from \(\mathcal{D}\), and
\begin{equation}
\rho_{j,t}(\theta)=
\frac{
\pi_{\theta}(o_{j,t}\mid q_j,o_{j,<t})
}{
\pi_{\theta_{\mathrm{old}}}(o_{j,t}\mid q_j,o_{j,<t})
}
\label{eq:grpo-ratio}
\end{equation}
is the token-level importance ratio, where $q_j$ denotes the prompt corresponding to response $o_j$.

\paragraph{Cross-model trajectory sharing.}
Cross-model RLVR allows heterogeneous models to learn from trajectories generated by their peers.
HACPO~\citep{zhang2026heterogeneous} broadly reuses peer rollouts during policy optimization, using capability-aware advantage estimation, sequence-level importance sampling, and clipping to account for cross-model mismatch. 
Its sharing is not restricted to prompts where the receiver's rollout group fails.
SGT~\citep{liu2026experience} instead transfers a verified peer success only when the receiver's entire rollout group fails and a peer succeeds, and learns from it through an auxiliary negative log-likelihood objective alongside on-policy GRPO.
These approaches raise two complementary design questions: \emph{which} peer trajectories should supplement the receiver's own rollouts, and \emph{how} the receiver should optimize on them under cross-model policy mismatch.

\section{GRAFT: Gated Replacement of Answer-Failed Groups with Peer Trajectories}
\label{sec:method}

\begin{figure}[t]
    \centering
    \includegraphics[width=\linewidth]{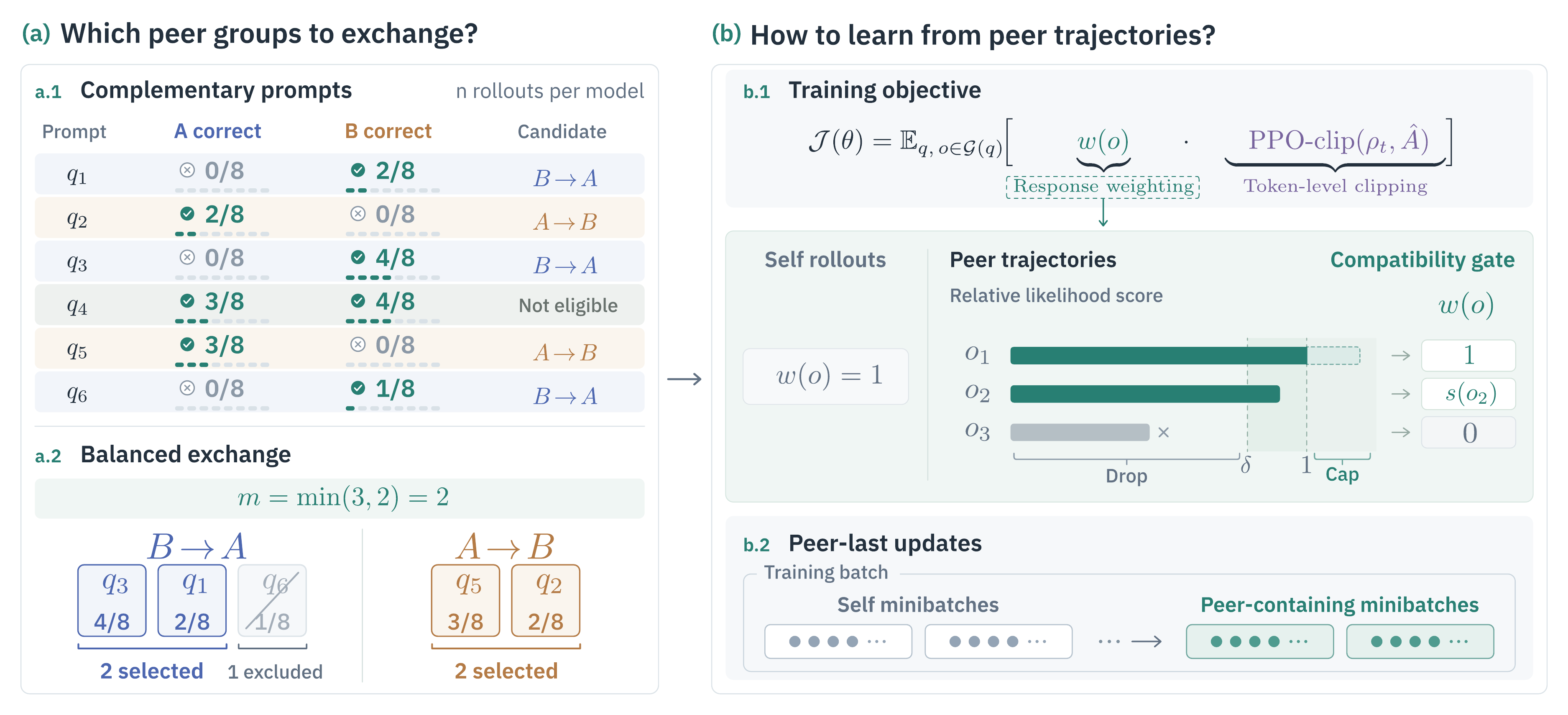}
    \caption{\textbf{Overall Framework of \method{}.} \method{} selects and balances complementary peer groups, then controls their off-policy influence through compatibility weighting, token-level clipping, and peer-last updates.}
    \label{fig:main}
\end{figure}

\paragraph{Overview.}
\method{} addresses the two design questions introduced in \Cref{sec:prelim}: \emph{which} peer trajectories to transfer, and \emph{how} the receiver should learn from them under cross-model mismatch (\Cref{fig:main}).
\method{} identifies complementary peer groups, replaces receiver groups that provide no successful trajectory, and balances transfer across the two directions (\Cref{sec:which}).
The receiver then learns from the transferred trajectories using source-computed advantages, sequence-level compatibility weighting, token-level importance ratio clipping, and peer-last updates (\Cref{sec:how}).

\paragraph{Setting.}
We consider two policies \(\pi_{\theta^A}\) and \(\pi_{\theta^B}\) trained simultaneously on the same prompt distribution \(\mathcal{D}\) with a verifiable binary reward.
The models maintain separate parameters and gradients, and exchange only sampled responses, their generation log-probabilities, and rewards.
A response $o$ is a string; $o^M=(o^M_1,\dots,o^M_{|o^M|})$ denotes its tokenization under $M$'s tokenizer, and for a policy $\pi$ on $M$'s vocabulary we set $\pi(o\mid q)=\prod_{t=1}^{|o^M|}\pi(o^M_t\mid q,o^M_{<t})$.\footnote{A superscript on a token sequence denotes the tokenizer; on an advantage, the model that computed it. String-level quantities ($r$, $s$, $w$) carry no superscript.}
For each prompt \(q\sim\mathcal{D}\), each model \(M\in\{A,B\}\) samples $\mathcal{G}_M(q)=(o_1,\dots,o_n)$ from its behavior policy \(\pi_{\theta^M_{\mathrm{old}}}\).
We denote the corresponding GRPO advantages by \(\{\hat{a}^M_i\}_{i=1}^n\), and define the number of successful responses as $k_M(q)=\sum_{o\in\mathcal{G}_M(q)} r(q,o)$.
Throughout, we describe transfer from model \(A\) to model \(B\), treating \(A\) as the source and \(B\) as the receiver; the reverse direction is symmetric.

\subsection{Complementary and Balanced Group Replacement}
\label{sec:which}

\paragraph{Complementary group replacement.}
Model $B$ receives peer trajectories only when its own rollout group fails entirely and the peer group contains both successful and unsuccessful responses:
\begin{equation}
\mathcal{C}_{A\to B}
=
\bigl\{
q \in \mathcal{Q}: k_B(q)=0 \;\wedge\; 1\leq k_A(q)<n
\bigr\},
\label{eq:gate}
\end{equation}
where $\mathcal{Q}\subset \mathcal{D}$. The condition $k_B(q)=0$ restricts transfer to prompts with no reward-based policy-gradient signal from the receiver's own group.
The condition $1\leq k_A(q)<n$ ensures that the peer group contains a verified success and nonzero reward variance.

For each prompt selected from $\mathcal{C}_{A\to B}$ by the balancing
procedure below, we replace $B$'s failed group with $A$'s entire rollout
group $\mathcal{G}_A(q)$; each transferred response enters $B$'s update
with its source-computed advantage $\hat{a}^{A}_i$ rather than a
re-normalized one.

Transferring both successful and unsuccessful responses preserves the reward contrast within the peer group, supplying positive and negative advantages without pooling rewards across models.
The receiver then applies the compatibility weights of \Cref{sec:how} while keeping the source advantages fixed.

\paragraph{Balanced exchange.}
Complementary candidate sets can differ substantially in size across directions, exposing one receiver to many more peer groups than the other.
We use the smaller candidate count,
$m=\min(|\mathcal{C}_{A\to B}|,|\mathcal{C}_{B\to A}|)$,
as a common selection target.
In each direction, candidates are ranked by the source model's success count in descending order, and the first $m$ prompts are retained together with all ties at the boundary.
This reduces directional imbalance while preserving equal-ranked candidates.
 
\subsection{Off-Policy-Aware Peer Updates}
\label{sec:how}

A grafted response is generated by $\pi_{\phi}\equiv\pi_{\theta^A_\mathrm{old}}$ and used to update $\pi_{\theta^B}$.
At the string level, the likelihood ratio factorizes as
\begin{equation}
\frac{\pi_{\theta^B}(o\mid q)}{\pi_{\phi}(o\mid q)}
=
\underbrace{
\frac{\pi_{\theta^B}(o^B\mid q)}
{\pi_{\theta^B_\mathrm{old}}(o^B\mid q)}
}_{\text{within-receiver change}}
\cdot
\underbrace{
\frac{\pi_{\theta^B_\mathrm{old}}(o^B\mid q)}
{\pi_{\phi}(o^A\mid q)}
}_{\text{cross-model mismatch}} ,
\label{eq:factorize}
\end{equation}
which separates the receiver's change during optimization from its initial mismatch with the peer.
We use this factorization to motivate treating the two discrepancies separately, with a token-level PPO surrogate for the former and a sequence-level compatibility weight for the latter; we do not use it to derive an exact importance-weighted objective.

To operationalize the cross-model mismatch under differing tokenizers, we define the average token log-likelihood
\begin{equation}
\bar{\ell}\bigl(\pi,o^M\mid q\bigr)
=
\frac{1}{|o^M|}\sum_{t=1}^{|o^M|}
\log \pi\bigl(o^M_t\mid q,o^M_{<t}\bigr).
\label{eq:avg-ll}
\end{equation}

\paragraph{Compatibility gate.}
We use sequence-level likelihood only to decide whether and how strongly a peer trajectory is admitted, and token-level importance ratio clipping to control the receiver's update on it.
We evaluate each tokenization under its corresponding model and define
\begin{equation}
s(o\mid q)
=
\exp\Bigl(
\bar{\ell}\bigl(\pi_{\theta^B_{\mathrm{old}}},o^B\mid q\bigr)
-
\bar{\ell}\bigl(\pi_{\phi},o^A\mid q\bigr)
\Bigr).
\label{eq:compat-score}
\end{equation}
This score compares average token log-likelihoods rather than accumulating log-probabilities over the entire response.
When the tokenizations coincide, \(s(o\mid q)\) reduces to the length-normalized sequence likelihood ratio; with different tokenizers, we instead interpret it as a compatibility score.

The score defines a bounded weight for each peer sequence,
\begin{equation}
w(o \mid q) = \mathbf{1}\big[s(o \mid q) > \delta\big]\cdot \min\{s(o \mid q),\, 1\}.
\label{eq:weight}
\end{equation}
When the prompt is clear from context, we abbreviate $w(o_j)\equiv w(o_j\mid q_j)$.
The threshold \(\delta\) excludes low-scoring peer trajectories regardless of
correctness: any response with $s(o\mid q)\le\delta$ is dropped from the
transferred group before optimization. The cap then limits the weight of each
admitted response to at most one. Correctness identifies a successful response, but on its own it does not say how well the receiver
can learn from that response under this update rule.

\paragraph{Token-level clipping.}
After group replacement, an optimization minibatch $\mathcal{B}$ may contain two kinds of
responses: self-generated responses $o_j\in\mathcal{G}_B(q_j)$ and grafted peer responses
$o_j\in\mathcal{G}_A(q_j)$.
Regardless of which model generated $o_j$, the receiver updates on its own tokenization
$o^{B}_j=(o^{B}_{j,1},\dots,o^{B}_{j,|o^{B}_j|})$; for grafted responses this amounts to
re-tokenizing the peer's response string with the receiver's tokenizer.
The token-level importance ratio is defined on this common receiver-side representation,
\begin{equation}
\rho_{j,t}(\theta^B)
=
\frac{\pi_{\theta^B}\bigl(o^{B}_{j,t}\mid q_j,\,o^{B}_{j,<t}\bigr)}
   {\pi_{\theta^B_{\mathrm{old}}}\bigl(o^{B}_{j,t}\mid q_j,\,o^{B}_{j,<t}\bigr)},
\label{eq:token-ratio}
\end{equation}
whose denominator is the receiver's behavior policy for both kinds of responses.
In particular, for a grafted response the denominator is \emph{not} the generating policy
$\pi_{\phi}$: following \Cref{eq:factorize}, the token-level ratio tracks only the
within-receiver change, while the cross-model mismatch is carried entirely by the
sequence-level weight.
The two kinds of responses therefore enter the objective with identically defined ratios and differ only in their weights and advantages: self-generated responses use $w(o_j)=1$ and $\hat{a}_j=\hat a^B_j$, while grafted responses use $w(o_j)$ from \Cref{eq:weight} and their source-computed $\hat{a}_j=\hat a^A_j$.

The receiver maximizes
\begin{equation}
\resizebox{0.92\linewidth}{!}{$\displaystyle
\mathcal{J}(\theta^B)
=
\mathbb{E}_{\mathcal{B}}\!\left[
\frac{1}{\sum_{j\in\mathcal{B}} |o^{B}_j|}
\sum_{j\in\mathcal{B}}
w(o_j)
\sum_{t=1}^{|o^{B}_j|}
\min\Bigl(
\rho_{j,t}(\theta^B)\hat{a}_j,\;
\operatorname{clip}
\bigl(\rho_{j,t}(\theta^B),1-\varepsilon_{\mathrm{low}},1+\varepsilon_{\mathrm{high}}\bigr)\hat{a}_j
\Bigr)
\right],
$}
\label{eq:objective}
\end{equation}
where both the advantages and the
compatibility weights are held fixed during receiver optimization.
Without transfer, every $w(o_j)=1$ and the objective reduces to the GRPO surrogate with
the same clipping settings.

\paragraph{Peer-last updates.}
Token-level clipping moderates peer contributions only once the importance ratios deviate from one. Before the first optimization step on a newly collected rollout batch, $\theta^B=\theta^B_\mathrm{old}$ and hence $\rho_{j,t}=1$, so a
grafted minibatch processed first would enter the update unclipped, and the sequence-level weight would be the only control on cross-model mismatch.
We therefore place minibatches containing grafted groups after the receiver's own on-policy minibatches.
By the time grafted responses are processed, clipping attenuates contributions whose ratios have moved outside $[1-\varepsilon_{\mathrm{low}},\,1+\varepsilon_{\mathrm{high}}]$ on the side determined by the sign of the advantage.
This ordering gives the receiver's own data priority and turns clipping into a second mechanism for moderating peer influence; we assess its empirical effect in \Cref{sec:ablation} and trace the resulting clipping dynamics in
Appendix~\ref{app:clip-dynamics}.

\section{Experiments}
\label{sec:exp}

\subsection{Setup}
\label{sec:exp-setup}

\paragraph{Models and pairs.}
We evaluate three heterogeneous model pairs: SmolLM3-3B-Base~\citep{bakouch2025smollm3}$\leftrightarrow$Qwen3-1.7B-Base~\citep{qwen3technicalreport} (Pair~1), OctoThinker-3B-Hybrid-Base~\citep{wang2025octothinker}$\leftrightarrow$Qwen3-1.7B-Base (Pair~2), and SmolLM3-3B-Base$\leftrightarrow$OctoThinker-3B-Hybrid-Base (Pair~3).
They differ in scale, tokenizer, pretraining corpus, and model architecture.

\paragraph{Training.}
All runs use \texttt{verl} with Ray, FSDP, and vLLM rollout. Each model samples $n{=}8$ responses per prompt.
Training data, learning rate, and training steps are held fixed across methods.
Unless stated otherwise, we use a compatibility gate threshold $\delta=0.8$ for all pairs.
Full hyperparameters are provided in Appendix~\ref{app:hparams}.

\paragraph{Evaluation.}
We report pass@1 on five mathematical benchmarks: MATH500~\citep{hendrycks2021measuring}, AIME2024, AIME2025, AMC23, and Minerva~\citep{lewkowycz2022solving}, together with their average.
Each checkpoint is evaluated over five runs with $8$ samples per prompt, and we report the mean and standard deviation.
$\Delta$ denotes the change in the five-benchmark average relative to GRPO ($n{=}8$).

\paragraph{Baselines.}
We compare against independent GRPO with $n{=}8$, $16$, and $32$ rollouts, as well as two cross-model training baselines, HACPO~\citep{zhang2026heterogeneous} and SGT~\citep{liu2026experience}.
\method{}, HACPO, and SGT all use $n{=}8$ rollouts per model, while GRPO with $n{=}16$ and $n{=}32$ gives a larger-rollout reference for assessing the benefit of additional independent exploration. Across methods, we keep the training data, learning rate, and number of training steps fixed. Appendix~\ref{app:hparams} lists the full baseline configurations.

\subsection{Main Results}
\label{sec:exp-main}

\begin{table*}[t]
\centering
\small
\renewcommand{\NC}{8}
\setlength{\tabcolsep}{5pt}
\caption{\textbf{Results of cross-model rollout exchange during co-training.}
Each updated policy uses the same GRPO baselines for both of its peer policies.
\textbf{Bold}: best among the $n{=}8$ methods [GRPO ($n{=}8$), HACPO, SGT, \method{}] within each peer block; no bold means GRPO ($n{=}8$) is best.
$^{\dagger}$/$^{\ddagger}$: \method{} exceeds all baselines up to GRPO ($n{=}16$)/($n{=}32$), respectively.
$\Delta$: change in average score relative to GRPO ($n{=}8$).}
\label{tab:main-results}
\resizebox{\textwidth}{!}{
\begin{tabular}{l c c c c c c c}
\toprule
\textbf{Method} & {MATH500} & {AIME2024} & {AIME2025} & {AMC23} & {Minerva} & {\textbf{Avg.}} & {$\Delta$\textsubscript{Avg}} \\
\midrule

\pairband{Updated policy: SmolLM3-3B-Base}
\quad GRPO ($n{=}8$)  & 72.08\stdv{0.52} & 8.58\stdv{1.13} & 8.50\stdv{0.96} & 46.62\stdv{1.49} & 27.22\stdv{0.63} & 32.60\stdv{0.49} & \refrow \\
\quad GRPO ($n{=}16$) & 72.88\stdv{0.40} & 9.08\stdv{0.90} & 9.83\stdv{1.34} & 47.19\stdv{1.01} & 27.90\stdv{0.75} & 33.38\stdv{0.25} & \up{0.78} \\
\quad GRPO ($n{=}32$) & 76.68\stdv{0.56} & 11.42\stdv{0.81} & 12.83\stdv{1.23} & 49.56\stdv{1.80} & 28.08\stdv{0.57} & 35.71\stdv{0.18} & \up{3.11} \\
\addlinespace[3pt]
\modelband{Peer policy: Qwen3-1.7B-Base (Pair 1)}
\quad HACPO & 69.58\stdv{0.53} & 7.08\stdv{1.79} & 10.83\stdv{1.69} & 42.88\stdv{1.61} & 27.17\stdv{0.61} & 31.51\stdv{0.43} & \dn{1.09} \\
\quad SGT & 75.53\stdv{0.51} & 10.67\stdv{1.52} & 11.50\stdv{0.37} & 49.06\stdv{1.51} & 27.63\stdv{0.68} & 34.88\stdv{0.21} & \up{2.28} \\
\oursrow \quad \textbf{\method{}} & \best{76.86$^{\ddagger}$\stdv{0.32}} & \best{14.42$^{\ddagger}$\stdv{1.49}} & \best{14.50$^{\ddagger}$\stdv{0.99}} & \best{50.87$^{\ddagger}$\stdv{1.60}} & \best{28.68$^{\ddagger}$\stdv{0.57}} & \best{37.06$^{\ddagger}$\stdv{0.67}} & \up{4.46} \\
\addlinespace[3pt]
\modelband{Peer policy: OctoThinker-3B-Hybrid-Base (Pair 3)}
\quad HACPO & 66.66\stdv{0.66} & 5.92\stdv{1.68} & 5.42\stdv{1.21} & 38.94\stdv{1.91} & 25.22\stdv{0.77} & 28.43\stdv{0.70} & \dn{4.17} \\
\quad SGT & 72.36\stdv{0.43} & 8.67\stdv{1.19} & 8.67\stdv{1.57} & 43.62\stdv{1.07} & 27.01\stdv{0.32} & 32.07\stdv{0.48} & \dn{0.53} \\
\oursrow \quad \textbf{\method{}} & \best{73.28$^{\dagger}$\stdv{0.14}} & \best{10.42$^{\dagger}$\stdv{1.02}} & \best{10.83$^{\dagger}$\stdv{0.78}} & 46.44\stdv{1.49} & 25.35\stdv{0.24} & \best{33.26\stdv{0.32}} & \up{0.66} \\
\midrule

\pairband{Updated policy: Qwen3-1.7B-Base}
\quad GRPO ($n{=}8$)  & 70.69\stdv{0.44} & 9.17\stdv{1.28} & 4.75\stdv{0.91} & 43.50\stdv{1.57} & 27.89\stdv{0.29} & 31.20\stdv{0.27} & \refrow \\
\quad GRPO ($n{=}16$) & 71.64\stdv{0.60} & 10.67\stdv{1.73} & 8.25\stdv{1.65} & 41.44\stdv{1.12} & 28.75\stdv{0.35} & 32.15\stdv{0.73} & \up{0.95} \\
\quad GRPO ($n{=}32$) & 71.90\stdv{0.10} & 10.17\stdv{0.56} & 5.75\stdv{0.75} & 45.62\stdv{2.31} & 28.71\stdv{0.56} & 32.43\stdv{0.65} & \up{1.23} \\
\addlinespace[3pt]
\modelband{Peer policy: SmolLM3-3B-Base (Pair 1)}
\quad HACPO & 63.74\stdv{0.60} & 7.17\stdv{0.90} & 4.17\stdv{1.18} & 36.06\stdv{1.20} & 25.04\stdv{0.44} & 27.24\stdv{0.30} & \dn{3.96} \\
\quad SGT & 70.97\stdv{0.77} & 10.17\stdv{1.34} & 5.92\stdv{0.85} & 40.75\stdv{0.87} & 28.39\stdv{0.43} & 31.24\stdv{0.56} & \up{0.04} \\
\oursrow \quad \textbf{\method{}} & \best{72.22$^{\ddagger}$\stdv{0.25}} & \best{12.50$^{\ddagger}$\stdv{0.42}} & \best{7.67\stdv{1.34}} & \best{45.56$^{\dagger}$\stdv{1.09}} & \best{29.23$^{\ddagger}$\stdv{0.32}} & \best{33.44$^{\ddagger}$\stdv{0.23}} & \up{2.24} \\
\addlinespace[3pt]
\modelband{Peer policy: OctoThinker-3B-Hybrid-Base (Pair 2)}
\quad HACPO & 66.93\stdv{0.30} & 6.83\stdv{0.96} & 4.00\stdv{0.48} & 36.69\stdv{1.78} & 27.64\stdv{0.38} & 28.42\stdv{0.38} & \dn{2.78} \\
\quad SGT & 69.47\stdv{0.25} & 9.58\stdv{0.29} & 5.83\stdv{1.28} & 41.69\stdv{1.90} & 28.42\stdv{0.81} & 31.00\stdv{0.40} & \dn{0.20} \\
\oursrow \quad \textbf{\method{}} & \best{71.75$^{\dagger}$\stdv{0.36}} & \best{11.33$^{\ddagger}$\stdv{0.99}} & \best{7.25\stdv{1.20}} & \best{44.56$^{\dagger}$\stdv{3.88}} & \best{29.02$^{\ddagger}$\stdv{0.77}} & \best{32.78$^{\ddagger}$\stdv{0.49}} & \up{1.58} \\
\midrule

\pairband{Updated policy: OctoThinker-3B-Hybrid-Base}
\quad GRPO ($n{=}8$)  & 56.32\stdv{0.60} & 2.50\stdv{0.93} & 1.08\stdv{0.86} & 27.94\stdv{1.75} & 18.06\stdv{0.81} & 21.18\stdv{0.40} & \refrow \\
\quad GRPO ($n{=}16$) & 58.39\stdv{0.69} & 3.25\stdv{0.54} & 1.83\stdv{0.48} & 28.69\stdv{0.87} & 19.86\stdv{0.50} & 22.40\stdv{0.43} & \up{1.22} \\
\quad GRPO ($n{=}32$) & 60.89\stdv{0.33} & 3.42\stdv{1.19} & 1.75\stdv{0.46} & 31.19\stdv{2.33} & 19.70\stdv{0.73} & 23.39\stdv{0.56} & \up{2.21} \\
\addlinespace[3pt]
\modelband{Peer policy: Qwen3-1.7B-Base (Pair 2)}
\quad HACPO & 55.44\stdv{0.51} & \best{3.58\stdv{1.49}} & 0.92\stdv{0.19} & 26.38\stdv{1.62} & 18.12\stdv{0.44} & 20.89\stdv{0.38} & \dn{0.29} \\
\quad SGT & \best{58.11\stdv{0.41}} & 2.17\stdv{1.26} & \best{2.08\stdv{0.59}} & 28.69\stdv{1.89} & 18.63\stdv{0.62} & 21.93\stdv{0.45} & \up{0.75} \\
\oursrow \quad \textbf{\method{}} & 58.01\stdv{0.36} & 3.42\stdv{0.80} & 1.25\stdv{0.78} & \best{34.75$^{\ddagger}$\stdv{2.98}} & \best{19.90$^{\ddagger}$\stdv{0.71}} & \best{23.47$^{\ddagger}$\stdv{0.68}} & \up{2.29} \\
\addlinespace[3pt]
\modelband{Peer policy: SmolLM3-3B-Base (Pair 3)}
\quad HACPO & 56.70\stdv{0.40} & 2.92\stdv{1.02} & 1.33\stdv{0.35} & 29.12\stdv{0.78} & 19.37\stdv{0.68} & 21.89\stdv{0.25} & \up{0.71} \\
\quad SGT & \best{57.82\stdv{0.31}} & 3.50\stdv{0.48} & 1.83\stdv{0.63} & 29.12\stdv{0.97} & \best{20.26\stdv{0.89}} & 22.51\stdv{0.32} & \up{1.33} \\
\oursrow \quad \textbf{\method{}} & 57.59\stdv{0.50} & \best{4.25$^{\ddagger}$\stdv{0.75}} & \best{2.25$^{\ddagger}$\stdv{0.70}} & \best{30.19$^{\dagger}$\stdv{1.30}} & 18.74\stdv{0.76} & \best{22.60$^{\dagger}$\stdv{0.22}} & \up{1.42} \\

\bottomrule
\end{tabular}
}
\end{table*}
 
\Cref{tab:main-results} reports the main comparison. \method{} improves over single-model GRPO in all six model blocks, with average-score gains between $+0.66$ and $+4.46$.
Rollout budget alone does not explain these gains.
On Pair~1, both models outperform GRPO with $4\times$ the rollouts ($n{=}32$), and on Pair~2 both are comparable to it.
Over three independent training runs, \method{} also has a higher mean aggregate score than budget-matched GRPO (Appendix~\ref{app:training-stability}).

The magnitude of improvement varies across pairs, with the largest gains on Pair~1 and the smallest on Pair~3.
Both co-training baselines are weaker.
HACPO falls below budget-matched GRPO in five of six blocks, by as much as $-4.17$ in average score. SGT does better but is inconsistent, ranging from $-0.53$ to $+2.28$. \method{}'s margin over the budget-matched baselines emerges early and persists throughout training rather than at an isolated checkpoint (\Cref{fig:val-curves}).

Qualitative case studies show shared solution steps between \method{} and peer responses, including root shifting and inclusion--exclusion, alongside elements of the receiver's GRPO solution structure (Appendix~\ref{app:graft-qualitative}).

\section{Analysis}
\label{sec:analysis}

We next examine three questions about the gains of \method{}: whether they come at a better compute trade-off than simply increasing the rollout budget (\Cref{sec:compute}), whether they strictly require synchronous co-training or persist with stored peer trajectories (\Cref{sec:stored}), and which components drive them, including whether the \emph{which} and \emph{how} decisions must be addressed together (\Cref{sec:ablation}).

\subsection{Compute-Efficient Gains from Peer Exchange}
\label{sec:compute}

\begin{figure}[t]
  \centering
  \begin{subfigure}[t]{0.48\textwidth}
    \includegraphics[width=\linewidth]{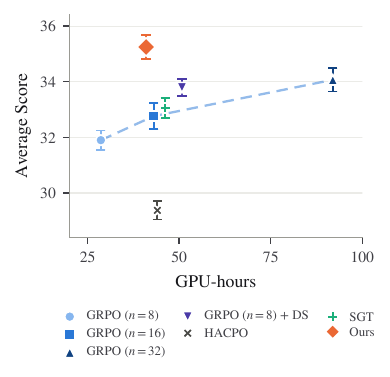}
    \caption{}\label{fig:pair-budget}
  \end{subfigure}
  \hfill
  \begin{subfigure}[t]{0.48\textwidth}
    \includegraphics[width=\linewidth]{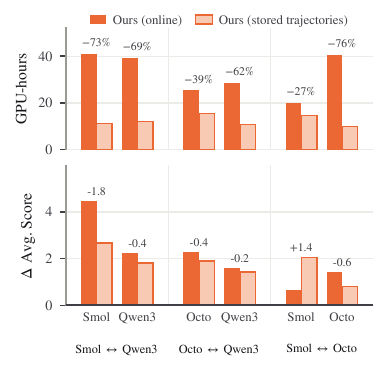}
    \caption{}\label{fig:stored-vs-online}
  \end{subfigure}
  \caption{\textbf{(a)} Average score of Pair 1 vs. total GPU-hours. Error bars: 95\% CIs over five evaluation runs; the dashed line traces GRPO with increasing rollout budget, and DS denotes dynamic sampling. GRAFT exceeds GRPO ($n{=}32$) by 1.18 at 0.45$\times$ the cost. \textbf{(b)} Replacing the co-trained partner with its stored trajectories across all six blocks. Top: GPU-hours to the reported checkpoint; bottom: gain over GRPO ($n{=}8$). Stored trajectories cut compute by
  27--76\% while keeping a positive gain in every block (84\% of the online gain on average).}
  \label{fig:pair-stored}
\end{figure}

We compare the GPU-hours required to obtain both models of Pair~1. 
Because \method{} jointly trains two policies, we compare the total cost of obtaining both resulting models rather than the cost of either model in isolation.
In \Cref{fig:pair-budget}, \method{} reaches a pair-mean score of $35.25$ at $40.9$ GPU-hours.
This exceeds GRPO ($n{=}32$) by $1.18$ points at $0.45\times$ its compute, and GRPO ($n{=}16$) by $2.48$ points at a comparable compute budget.
Thus, increasing independent rollout budgets does not match the performance-compute trade-off of peer exchange in this comparison.
Appendix~\ref{app:compute} defines checkpoint-cost accounting and reports full-run costs and per-model results.

\subsection{Using Stored Peer Trajectories during Training}
\label{sec:stored}

Online co-training keeps both models and their optimizer states resident. We therefore test whether \method{} can instead use peer trajectories stored from the partner's independent GRPO ($n{=}8$) run: at each receiver step we load the partner's recorded responses and log-probabilities for the same prompt batch, re-verify their rewards, and apply the same selection, weighting, and update rule (Appendix~\ref{app:hparams}). Transfer is unidirectional, so balanced exchange is inactive.

Stored trajectories improve over GRPO ($n{=}8$) in all six model blocks, by $1.78$ points on average compared with $2.11$ for online exchange (\Cref{fig:stored-vs-online}; full results in Appendix~\ref{app:full-stored}).
On Pair~1, obtaining both selected receiver checkpoints requires $23.2$ GPU-hours, excluding the prior GRPO runs used to collect the peer logs.
Most of the performance benefit therefore persists when existing peer trajectories are reused, without simultaneous co-training or keeping the peer model in memory.

\subsection{Ablations and Alternative Designs}
\label{sec:ablation}

\begin{table}[t]
\centering
\small
\setlength{\tabcolsep}{4pt}
\caption{\textbf{Ablations and alternative designs on Pair~1}
(S: SmolLM3-3B, Q: Qwen3-1.7B; $\Delta$: change vs.\ full \method{}).
(a) removes or randomizes one component at a time;
(b) replaces the transfer rule, or fixes only one of \emph{which} and
\emph{how} while borrowing the other.
Variant definitions and per-benchmark scores: Appendices~\ref{app:alt-designs} and \ref{app:ablation-full}.}
\label{tab:ablation}
\begin{subtable}[t]{0.495\linewidth}
\centering
\caption{Component ablation}
\label{tab:ablation-comp}
\resizebox{\linewidth}{!}{
\begin{tabular}{l cc cc}
\toprule
& \multicolumn{2}{c}{\textbf{S}} & \multicolumn{2}{c}{\textbf{Q}} \\
\cmidrule(lr){2-3} \cmidrule(lr){4-5}
& Avg. & $\Delta$ & Avg. & $\Delta$ \\
\midrule
\oursrow \method{} (full) & 37.06\stdv{0.67} & \refrow & 33.44\stdv{0.23} & \refrow \\
\midrule
\multicolumn{5}{l}{\emph{Component removal}} \\
\quad compat.\ gate & 28.70\stdv{0.78} & \dn{8.36} & 30.93\stdv{0.52} & \dn{2.51} \\
\quad compat.\ floor & 35.17\stdv{0.59} & \dn{1.89} & 29.88\stdv{0.62} & \dn{3.56} \\
\quad balancing & 35.06\stdv{0.27} & \dn{2.00} & 32.96\stdv{0.56} & \dn{0.48} \\
\quad token ratio & 35.21\stdv{0.65} & \dn{1.85} & 32.99\stdv{0.33} & \dn{0.45} \\
\midrule
\multicolumn{5}{l}{\emph{Random selection}} \\
\quad prompts & 34.66\stdv{0.47} & \dn{2.40} & 31.86\stdv{0.48} & \dn{1.58} \\
\quad admission & 35.20\stdv{0.85} & \dn{1.86} & 32.35\stdv{0.32} & \dn{1.09} \\
\midrule
\multicolumn{5}{l}{\emph{Peer position}} \\
\quad first & 34.61\stdv{0.43} & \dn{2.45} & 31.73\stdv{0.58} & \dn{1.71} \\
\quad uniform & 33.49\stdv{0.47} & \dn{3.57} & 32.41\stdv{0.55} & \dn{1.03} \\
\midrule
GRPO & 32.60\stdv{0.49} & \dn{4.46} & 31.20\stdv{0.27} & \dn{2.24} \\
\bottomrule
\end{tabular}}
\end{subtable}\hfill
\begin{subtable}[t]{0.495\linewidth}
\centering
\caption{Alternative designs}
\label{tab:ablation-design}
\resizebox{\linewidth}{!}{
\begin{tabular}{l cc cc}
\toprule
& \multicolumn{2}{c}{\textbf{S}} & \multicolumn{2}{c}{\textbf{Q}} \\
\cmidrule(lr){2-3} \cmidrule(lr){4-5}
& Avg. & $\Delta$ & Avg. & $\Delta$ \\
\midrule
\oursrow \method{} (full) & 37.06\stdv{0.67} & \refrow & 33.44\stdv{0.23} & \refrow \\
\midrule
\multicolumn{5}{l}{\emph{Transfer rule}} \\
\quad pooled groups & 33.07\stdv{0.47} & \dn{3.99} & 30.78\stdv{0.48} & \dn{2.66} \\
\quad success-only & 29.22\stdv{0.26} & \dn{7.84} & 29.51\stdv{0.65} & \dn{3.93} \\
\quad \; $+$ w/o floor & 32.14\stdv{0.38} & \dn{4.92} & 29.85\stdv{0.44} & \dn{3.59} \\
\midrule
\multicolumn{5}{l}{\emph{Our which, prior how}} \\
\quad HACPO update & 28.97\stdv{0.52} & \dn{8.09} & 27.79\stdv{0.38} & \dn{5.65} \\
\quad SFT update (SGT) & 34.08\stdv{0.15} & \dn{2.98} & 31.89\stdv{0.37} & \dn{1.55} \\
\quad LUFFY update & 32.51\stdv{0.89} & \dn{4.55} & 31.41\stdv{0.50} & \dn{2.03} \\
\midrule
\multicolumn{5}{l}{\emph{Original methods}} \\
\quad HACPO & 31.51\stdv{0.43} & \dn{5.55} & 27.24\stdv{0.30} & \dn{6.20} \\
\quad SGT & 34.88\stdv{0.21} & \dn{2.18} & 31.24\stdv{0.56} & \dn{2.20} \\
\midrule
GRPO & 32.60\stdv{0.49} & \dn{4.46} & 31.20\stdv{0.27} & \dn{2.24} \\
\bottomrule
\end{tabular}}
\end{subtable}
\end{table}

\Cref{tab:ablation-comp} evaluates individual design choices on Pair~1.
Removing compatibility weighting causes the largest degradation for SmolLM3 ($-8.36$ points), while removing the floor causes the largest degradation for Qwen3 ($-3.56$).
Removing balancing or replacing the token-level ratio with a sequence-level ratio also lowers performance.
Count-matched random prompts and random admission also underperform.
Peer-last ordering beats peer-first and uniform; Appendix~\ref{app:clip-dynamics} additionally shows that it yields the highest clipping rate on peer tokens.
We use $\delta=0.8$ throughout, with a threshold sweep in Appendix~\ref{app:ablation-full}.

\Cref{tab:ablation-design} tests alternative designs.
Pooling peer and self groups or transferring only successes underperforms full \method{}, and the floor improves aggregate scores only under full-group transfer.
With our prompt selection fixed, HACPO, SGT, and LUFFY-style~\citep{NEURIPS2025_a9d5c33e} updates all trail \method{}.
Using \method{}'s update with SGT's unbalanced selection also leaves a gap of 2.00/0.48 points.
Together, these comparisons support combining complementary and balanced prompt selection with compatibility-aware peer updates.

\section{Conclusion}
\label{sec:conclusion}

We have introduced \textbf{\method{}}, an off-policy-aware framework for cross-model trajectory exchange in RLVR. \method{} exploits complementary successes across heterogeneous models by replacing all-fail rollout groups with informative peer groups, while controlling cross-model mismatch through compatibility-aware and clipped updates. Across three heterogeneous model pairs, \method{} consistently improves both models over standard GRPO with the same rollout budget and can match or exceed GRPO with substantially more rollouts. Moreover, most of the gains persist when using stored peer trajectories, showing that the benefit of cross-model exploration does not require synchronous co-training. 

\paragraph{Limitations.} 
\method{}'s gain depends on how complementary the two models are and is smallest on Pair 3. Across tokenizers, the compatibility score is a proxy, not a density ratio. We also study only two-model pairs, only on math, and only with base models of at most 3B parameters. We leave exchange among more than two peers, and in domains without verifiable rewards, to future work.

\bibliographystyle{preprint}
\bibliography{references}

\newpage
\appendix

\section{Full Algorithm of \method{}}
\label{app:full-algo}

\begin{algorithm}[H]
\caption{GRAFT: Cross-Model Trajectory Exchange}
\label{alg:graft}
\begin{algorithmic}[1]
\Require Policies $\pi_{\theta^A},\pi_{\theta^B}$;
         prompt set $\mathcal{D}$; rollout count $n$;
         compatibility threshold $\delta$;
         clipping parameters
         $\varepsilon_{\mathrm{low}},\varepsilon_{\mathrm{high}}$

\For{each training step}
    \State Sample a shared prompt batch $\mathcal{Q}\subset\mathcal{D}$
    \For{$M\in\{A,B\}$}
        \State $\theta^M_{\mathrm{old}}\gets\theta^M$
        \State Sample $n$ responses $\mathcal{G}_M(q)$ from
               $\pi_{\theta^M_{\mathrm{old}}}$ for each $q\in\mathcal{Q}$
        \State Store source tokens and generation log-probabilities
        \State Compute rewards, success counts $k_M(q)$,
               and advantages $\hat{a}_M(q)$
        \State Set advantages to zero for zero-variance groups
    \EndFor

    \Statex \Comment{Select complementary groups with nonzero reward variance}
    \For{$(S,R)\in\{(A,B),(B,A)\}$}
        \State $\mathcal{C}_{S\to R}\gets
               \{q\in\mathcal{Q}:k_R(q)=0,\ 1\leq k_S(q)<n\}$
    \EndFor
    \State $m\gets
           \min(|\mathcal{C}_{A\to B}|,|\mathcal{C}_{B\to A}|)$
    \For{$(S,R)\in\{(A,B),(B,A)\}$}
        \State $\mathcal{E}_{S\to R}\gets
               \Call{SelectWithTies}{\mathcal{C}_{S\to R},k_S,m}$
    \EndFor

    \Statex \Comment{Replace selected groups using the original source rollouts}
    \For{$(S,R)\in\{(A,B),(B,A)\}$}
        \State Initialize receiver training buffer $\mathcal{B}_R\gets\varnothing$
        \For{$q\in\mathcal{Q}$}
            \If{$q\in\mathcal{E}_{S\to R}$}
                \State Re-tokenize $\mathcal{G}_S(q)$ for receiver $R$
                \State Compute $s(o\mid q)$ using
                       $\pi_{\theta^R_{\mathrm{old}}}$ and stored
                       source log-probabilities
                \State $w(o)\gets
                       \mathbf{1}[s(o\mid q)>\delta]\min\{s(o\mid q),1\}$
                \State Add the admitted peer responses ($s(o\mid q)>\delta$) with source advantages $\hat{a}_S(q)$ and weights $w(o)$ to $\mathcal{B}_R$
            \Else
                \State Add $\mathcal{G}_R(q)$ with advantages
                       $\hat{a}_R(q)$ and unit weights to $\mathcal{B}_R$
            \EndIf
        \EndFor
    \EndFor

    \Statex \Comment{Optimize with fixed advantages and compatibility weights}
    \For{$R\in\{A,B\}$}
        \For{each optimization epoch}
            \State Arrange $\mathcal{B}_R$ into minibatches,
                   placing peer-containing minibatches last
            \For{each minibatch in this order}
                \State Compute token ratios relative to
                       $\pi_{\theta^R_{\mathrm{old}}}$
                \State Update $\theta^R$ by gradient ascent on \Cref{eq:objective}
            \EndFor
        \EndFor
    \EndFor
\EndFor
\State \Return $\pi_{\theta^A},\pi_{\theta^B}$
\end{algorithmic}
\end{algorithm}

$\operatorname{SelectWithTies}(\mathcal{C},k_S,m)$ ranks candidate prompts by the source model's success count $k_S(q)$ in descending order and retains the first $m$ prompts, including all ties at the boundary. It returns $\varnothing$ when $m=0$. The selected counts may exceed $m$ and need not be identical across directions. Compatibility weights and group advantages remain fixed during optimization. Peer responses failing the compatibility floor are removed before optimization; group advantages are computed on the full source group prior to this removal and remain fixed.

\section{Training and Evaluation Details}
\label{app:hparams}

\paragraph{Data and reward.}
We train on the 7,500 problems in the MATH training split, using all difficulty levels. Each problem is formatted as a single user message with the suffix \textit{``Let's think step by step and output the final answer within \textbackslash boxed\{\}.''} The reward is binary: \texttt{math\_verify} checks the final boxed answer against the reference answer, with verification timeouts assigned zero reward. We use no additional format or length reward, and score responses that reach the generation limit as generated.

\paragraph{Optimization and implementation.}
We use \texttt{verl} with FSDP for policy optimization and vLLM~0.8.5 for rollout generation. Each pair is trained on four NVIDIA H200 GPUs, with both models colocated on the same node and rollout tensor parallelism set to two. Table~\ref{tab:training-hparams} summarizes the common configuration; baseline-specific exceptions are described below. Both models receive the same prompt batch at each training step. We perform one optimization epoch per rollout batch, partitioned into four minibatches of 32 prompts. The implementation averages the policy loss over response tokens (\texttt{token-mean} aggregation). Training uses dynamic microbatching and gradient checkpointing.

\begin{table}[H]
\centering
\small
\setlength{\tabcolsep}{8pt}
\caption{\textbf{Training hyperparameters.}
These settings apply to \method{} and independent GRPO unless otherwise
specified. The larger-budget GRPO baselines change only the rollout count.}
\label{tab:training-hparams}
\begin{tabular}{@{}ll@{}}
\toprule
Hyperparameter & Value \\
\midrule
Training epochs& 3 \\
Prompt batch size & 128 \\
Rollouts per prompt per model & 8 \\
Optimization minibatch size & 32 prompts \\
Optimization epochs per rollout batch & 1 \\
Maximum prompt / response length & 2,048 / 4,096 tokens \\
Rollout temperature / top-$p$ & 1.0 / 1.0 \\
Optimizer & AdamW \\
Learning rate & $10^{-6}$, constant, no warmup \\
Adam coefficients $(\beta_1,\beta_2)$ & $(0.9,0.999)$ \\
Weight decay & 0.01 \\
Maximum gradient norm & 1.0 \\
PPO clipping $(\varepsilon_{\mathrm{low}},\varepsilon_{\mathrm{high}})$
& $(0.2,0.28)$ \\
GRPO normalization stabilizer & $10^{-6}$ \\
KL regularization / entropy bonus & none / none \\
Precision & bfloat16 \\
\bottomrule
\end{tabular}
\end{table}

\paragraph{GRAFT configuration.}
We use the same exchange and optimization settings for all three pairs. The compatibility threshold $\delta=0.8$ was selected on Pair~1 and fixed
before running Pairs~2 and~3. Exchange is performed at every training step using the selection and balancing rules in \Cref{sec:which}, with no additional exchange-volume cap. Peer responses are re-tokenized for the receiver, while source-computed advantages and compatibility weights remain fixed throughout optimization. Minibatches containing peer groups are placed last. The method uses no additional rollouts; its additional computation is receiver-side likelihood evaluation of transferred responses.

\paragraph{Baseline configurations.}
Independent GRPO uses the same training settings with $n\in\{8,16,32\}$ and no trajectory exchange. The minibatch size remains fixed at 32 prompts, so larger rollout groups increase the number of responses per update while preserving the number of optimizer updates. For HACPO~\citep{zhang2026heterogeneous}, we retain the released method configuration: sequence-level clipping with lower and upper clip deltas of $3\times10^{-4}$ and $4\times10^{-4}$, a peer-rollout clipping lower bound initialized at 0.8 and increased by 0.025 per step, a peer loss coefficient of 1.0, a minibatch size of 64 prompts, and a KL loss coefficient of $10^{-3}$, following the default configuration in their official implementation. HACPO pools peer rollouts for every prompt and uses $n{=}8$, the same learning rate, and the same training duration. We also tested HACPO without KL regularization and with 32-prompt minibatches; the reported configuration performed better. For SGT~\citep{liu2026experience}, we augment the GRPO loss with a supervised negative log-likelihood term weighted by $\lambda=0.1$, following the paper's original setting. Whenever a receiver has no correct rollout and its peer has at least one, we uniformly sample one correct peer response for this term. This rule is applied in both directions. Variants that combine a baseline with a component of \method{} are described in Appendix~\ref{app:alt-designs}.

\paragraph{Saved peer logs.}
For experiments using saved peer logs, the trajectories come from the partner's independent GRPO ($n{=}8$) training run. At each receiver training step, we load the corresponding recorded peer responses and generation log-probabilities, re-evaluate their rewards, and apply the same selection, compatibility weighting, and receiver update. The partner model is not loaded or trained. Directional balancing is inactive because transfer is unidirectional.

\paragraph{Evaluation.}
We evaluate on MATH500 (500 problems), AIME2024 (30), AIME2025 (30), AMC23 (40), and Minerva Math (272), using the same prompt formatting, answer verifier, and length limits as in training, with temperature 0.6, top-$p$ 0.95, and eight responses per problem.
We estimate pass@1 by averaging correctness over responses and then over problems; the aggregate score is the unweighted mean of the five benchmarks.
Since RLVR training curves are non-monotonic across nearby checkpoints, it is common to monitor training on validation sets drawn from the evaluation benchmarks~\citep{dapo,liu2025prorl} and report the best checkpoint so selected~\citep{le2026no,kim2026discounted,jiang2025rethinkingentropyregularizationlarge}.

\begin{figure}[H]
  \centering
  \includegraphics[width=\linewidth]{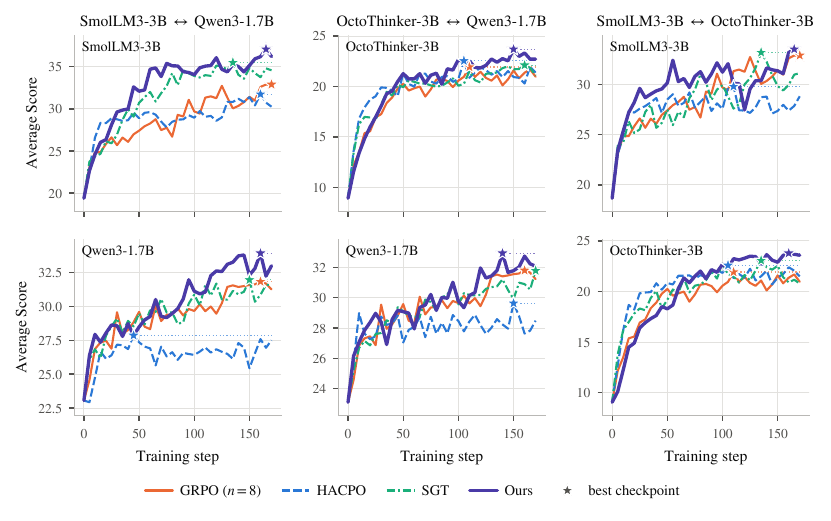}
  \caption{\textbf{Validation trajectories used for checkpoint selection.}
  Five-benchmark average vs.\ training step for each updated model,
  evaluated under the identical protocol for all methods.
  Stars mark the selected checkpoints in \Cref{tab:main-results}.}
  \label{fig:val-curves}
\end{figure}

We follow this practice with a single fixed rule: every model is validated every five steps on the aggregate score and the highest-scoring checkpoint is reported, under the identical rule for every method and rollout budget.
\Cref{fig:val-curves} shows that \method{} leads the budget-matched baselines throughout most of training.
Each selected checkpoint is evaluated over five independent runs, and we report their mean and standard deviation.
For Appendix~\ref{app:training-stability}, we average the five evaluation scores within each training run and report the mean and sample standard deviation over three independent runs; main-table results use the first run.

\clearpage

\section{Training Stability Across Independent Runs}
\label{app:training-stability}

To assess sensitivity to training variability, we independently repeat both GRPO ($n{=}8$) and \method{} three times for all three model pairs. For each training run, we evaluate the resulting checkpoint over five evaluation runs with $8$ samples per prompt and report their mean performance. Table~\ref{tab:training-variance} reports the mean and sample standard deviation across the three independent training runs. The main results in Table~\ref{tab:main-results} use the first training run for each setting. Across all six model blocks, \method{} maintains higher mean aggregate performance than the single-model GRPO baseline, indicating that the improvements persist across independent training runs.

\begin{table}[H]
\centering
\small
\setlength{\tabcolsep}{5pt}
\caption{\textbf{Training stability across three independent runs.}
We repeat GRPO ($n{=}8$) and \method{} three times for all three model pairs and report the mean and sample standard deviation across runs.
$\Delta$ denotes the difference in the mean average score relative to GRPO ($n{=}8$) within each model block.
\textbf{Bold} indicates the better mean performance between GRPO ($n{=}8$) and \method{}.}
\label{tab:training-variance}
\resizebox{\textwidth}{!}{
\begin{tabular}{l c c c c c c c}
\toprule
\textbf{Method}
& {MATH500}
& {AIME2024}
& {AIME2025}
& {AMC23}
& {Minerva}
& {\textbf{Avg.}}
& {$\Delta$\textsubscript{Avg}} \\
\midrule

\pairband{Pair 1\quad SmolLM3-3B-Base $\leftrightarrow$ Qwen3-1.7B-Base}

\modelband{SmolLM3-3B-Base}
\quad GRPO ($n{=}8$)
& 72.21\stdv{2.78}
& 7.91\stdv{0.58}
& 8.72\stdv{2.01}
& 45.08\stdv{2.46}
& 26.84\stdv{0.45}
& 32.15\stdv{1.48}
& \refrow \\
\oursrow \quad \method{}
& \best{75.37\stdv{1.30}}
& \best{13.28\stdv{1.32}}
& \best{13.22\stdv{1.29}}
& \best{50.37\stdv{0.98}}
& \best{28.34\stdv{1.36}}
& \best{36.12\stdv{0.85}}
& \up{3.97} \\

\addlinespace[3pt]
\modelband{Qwen3-1.7B-Base}
\quad GRPO ($n{=}8$)
& 70.69\stdv{0.30}
& 9.66\stdv{0.43}
& 5.86\stdv{0.97}
& 42.60\stdv{0.78}
& 28.10\stdv{0.35}
& 31.38\stdv{0.16}
& \refrow \\
\oursrow \quad \method{}
& \best{71.78\stdv{0.38}}
& \best{12.06\stdv{0.57}}
& \best{7.33\stdv{0.96}}
& \best{45.81\stdv{0.29}}
& \best{29.22\stdv{0.09}}
& \best{33.24\stdv{0.33}}
& \up{1.86} \\

\midrule

\pairband{Pair 2\quad OctoThinker-3B-Hybrid-Base $\leftrightarrow$ Qwen3-1.7B-Base}

\modelband{OctoThinker-3B-Hybrid-Base}
\quad GRPO ($n{=}8$)
& 55.86\stdv{0.42}
& 2.94\stdv{1.15}
& 0.72\stdv{0.38}
& 28.35\stdv{0.36}
& 17.82\stdv{0.40}
& 21.14\stdv{0.22}
& \refrow \\
\oursrow \quad \method{}
& \best{57.36\stdv{0.57}}
& \best{3.67\stdv{0.90}}
& \best{2.05\stdv{0.94}}
& \best{33.56\stdv{2.11}}
& \best{19.30\stdv{0.72}}
& \best{23.19\stdv{0.24}}
& \up{2.05} \\

\addlinespace[3pt]
\modelband{Qwen3-1.7B-Base}
\quad GRPO ($n{=}8$)
& 70.69\stdv{0.30}
& 9.66\stdv{0.43}
& 5.86\stdv{0.97}
& 42.60\stdv{0.78}
& 28.10\stdv{0.35}
& 31.38\stdv{0.16}
& \refrow \\
\oursrow \quad \method{}
& \best{71.17\stdv{0.53}}
& \best{11.19\stdv{0.64}}
& \best{7.72\stdv{0.97}}
& \best{43.19\stdv{1.44}}
& \best{28.57\stdv{0.48}}
& \best{32.37\stdv{0.41}}
& \up{0.99} \\
\midrule

\pairband{Pair 3\quad SmolLM3-3B-Base $\leftrightarrow$ OctoThinker-3B-Hybrid-Base}

\modelband{SmolLM3-3B-Base}
\quad GRPO ($n{=}8$)
& 72.21\stdv{2.78}
& 7.91\stdv{0.58}
& 8.72\stdv{2.01}
& 45.08\stdv{2.46}
& \best{26.84\stdv{0.45}}
& 32.15\stdv{1.48}
& \refrow \\
\oursrow \quad \method{}
& \best{72.27\stdv{1.13}}
& \best{9.92\stdv{1.96}}
& \best{10.19\stdv{0.90}}
& \best{46.92\stdv{0.94}}
& 26.03\stdv{0.63}
& \best{33.06\stdv{0.31}}
& \up{0.91} \\

\addlinespace[3pt]
\modelband{OctoThinker-3B-Hybrid-Base}
\quad GRPO ($n{=}8$)
& 55.86\stdv{0.42}
& 2.94\stdv{1.15}
& 0.72\stdv{0.38}
& 28.35\stdv{0.36}
& 17.82\stdv{0.40}
& 21.14\stdv{0.22}
& \refrow \\
\oursrow \quad \method{}
& \best{56.47\stdv{2.47}}
& \best{5.03\stdv{1.35}}
& \best{1.75\stdv{0.60}}
& \best{30.25\stdv{0.06}}
& \best{19.29\stdv{1.57}}
& \best{22.56\stdv{1.07}}
& \up{1.42} \\

\bottomrule
\end{tabular}
}
\end{table}

\clearpage
\section{Compute Accounting}
\label{app:compute}

\begin{figure}[H]
  \centering
  \includegraphics[width=\textwidth]{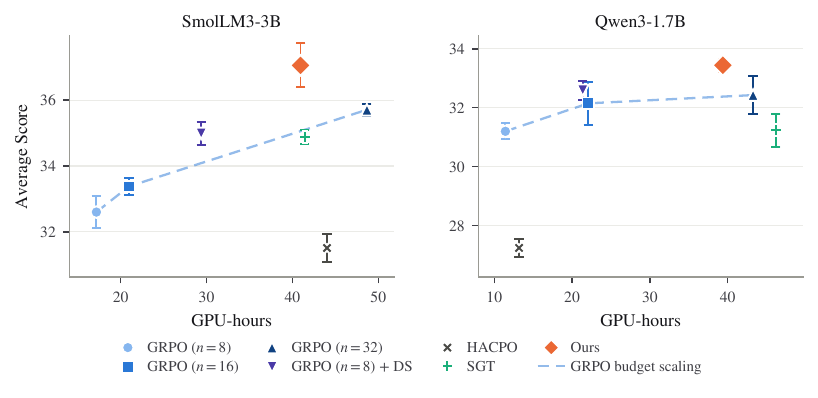}
  \caption{Per-model score against the GPU-hours charged to that model under the most conservative rule: GRPO pays for its own model only; each model of a
  cross-model method is charged the full joint-run cost (both models' training) up
  to that model's own selected checkpoint. Error bars are $\pm 1$ s.d.\ over inference seeds.}
  \label{fig:compute-mixed}
\end{figure}

\paragraph{Measurement.}
GPU-hours are computed as $N_{\mathrm{GPU}}$ times the summed per-step wall-clock training time, excluding validation and checkpoint writing.
All runs in \Cref{tab:compute} use four GPUs.
We report both the cost of obtaining the selected checkpoints and the cost of completing all three training epochs (174 steps).

For independent GRPO, the cost of obtaining both models is the sum of the costs of their separate runs up to their respective selected checkpoints.
For online cross-model methods, the two selected checkpoints can occur at different steps.
The joint checkpoint cost is therefore the cost of running the joint training process through the later of these two steps.
It is not generally equal to the sum of the two per-model checkpoint entries.
The full-run cost reports the complete training budget without assuming that the selected checkpoint is known in advance.

\paragraph{Conservative per-model accounting.}
\Cref{fig:compute-mixed} charges independent GRPO only for the model it
produces, but charges each model of an online cross-model method the full cost
of the joint run, covering both models' training, up to that model's own
selected checkpoint (e.g., $40.9$ GPU-hours for SmolLM3 at step 165 and $39.4$
for Qwen3 at step 160 under \method{}).
This differs from the joint checkpoint cost in \Cref{tab:compute}, which runs
through the later of the two checkpoints.
Under this accounting, \method{} exceeds GRPO ($n{=}32$) by $1.35$ points for
SmolLM3 and $1.01$ points for Qwen3, at approximately $0.84\times$ and
$0.91\times$ their respective checkpoint costs.
Compared with GRPO ($n{=}16$), it improves scores by $3.68$ and $1.29$ points at
approximately $1.95\times$ and $1.78\times$ the cost.
Thus, the advantage over the largest tested rollout budget persists even when
the full joint cost is charged to each model separately.

\begin{table}[t]
\centering
\small
\setlength{\tabcolsep}{6pt}
\caption{\textbf{Score and training cost on Pair~1.}
Avg.\ is the five-benchmark average at the selected checkpoint (\Cref{tab:main-results}).
GPU-hour columns report the cost up to the selected checkpoint and for the complete three-epoch run.
For online cross-model methods, the ``Both models / Best ckpt'' entry measures the joint run through the later of the two selected checkpoints; for independent GRPO, it sums the two separate checkpoint costs.
Stored-trajectory costs exclude the prior runs used to collect peer logs.}
\label{tab:compute}
\resizebox{\textwidth}{!}{
\begin{tabular}{l ccc ccc cc}
\toprule
& \multicolumn{3}{c}{\textbf{SmolLM3-3B-Base}} & \multicolumn{3}{c}{\textbf{Qwen3-1.7B-Base}}
& \multicolumn{2}{c}{\textbf{Both models}} \\
\cmidrule(lr){2-4} \cmidrule(lr){5-7} \cmidrule(lr){8-9}
& Avg. & Best ckpt & 3 epochs & Avg. & Best ckpt & 3 epochs & Best ckpt & 3 epochs \\
& (\%) & (GPU-h) & (GPU-h) & (\%) & (GPU-h) & (GPU-h) & (GPU-h) & (GPU-h) \\
\midrule
GRPO ($n{=}8$) & 32.60\stdv{0.49} & 17.2 & 17.4 & 31.20\stdv{0.27} & 11.4 & 12.5 & 28.6 & 29.9 \\
GRPO ($n{=}16$) & 33.38\stdv{0.25} & 21.0 & 34.8 & 32.15\stdv{0.73} & 22.1 & 22.5 & 43.1 & 57.3 \\
GRPO ($n{=}32$) & 35.71\stdv{0.18} & 48.7 & 99.5 & 32.43\stdv{0.65} & 43.3 & 45.6 & 92.0 & 145.1 \\
GRPO ($n{=}8$) + Dynamic Sampling & 34.99\stdv{0.35} & 29.4 & 40.5 & 32.59\stdv{0.33} & 21.4 & 27.4 & 50.8 & 67.9 \\
HACPO & 31.51\stdv{0.43} & 23.5 & 25.1 & 27.24\stdv{0.30} & 5.9 & 22.0 & 44.0 & 47.1 \\
SGT & 34.88\stdv{0.21} & 24.0 & 30.9 & 31.24\stdv{0.56} & 19.4 & 22.2 & 46.2 & 53.1 \\
\oursrow \method{} & 37.06\stdv{0.67} & 25.3 & 26.7 & 33.44\stdv{0.23} & 15.0 & 16.6 & 40.9 & 43.3 \\
\oursrow \method{} w/ stored traj. & 35.28\stdv{0.33} & 11.1 & 20.0 & 33.01\stdv{0.53} & 12.1 & 13.3 & 23.2 & 33.3 \\
\bottomrule
\end{tabular}
}
\end{table}

\clearpage

\section{Full Results for Stored Trajectory Experiments}
\label{app:full-stored}

Table~\ref{tab:replay-results} reports per-benchmark results for the stored trajectory experiments in \Cref{sec:stored}.
\method{} with stored trajectories improves the average score over GRPO ($n{=}8$) in all six model blocks, by $0.80$--$2.68$ points ($1.78$ on average, 84\% of the $2.11$ point online gain), and exceeds online \method{} for SmolLM3-3B-Base in Pair~3.

\begin{table}[H]
\centering
\small
\renewcommand{\NC}{8}
\setlength{\tabcolsep}{5pt}
\caption{\textbf{Effect of learning from peer-model training trajectories.}
We compare standard GRPO ($n{=}8$), online cross-model rollout exchange (\method{}), and \method{} with peer replay, where each model learns from peer trajectories collected from a previous training run rather than from a simultaneously co-trained peer.
All methods use $n{=}8$ self-rollouts per model.
$\Delta$ reports the change in average score relative to GRPO ($n{=}8$) within each model block.}
\label{tab:replay-results}
\resizebox{\textwidth}{!}{
\begin{tabular}{l c c c c c c c}
\toprule
\textbf{Method} & {MATH500} & {AIME2024} & {AIME2025} & {AMC23} & {Minerva} & {\textbf{Avg.}} & {$\Delta$\textsubscript{Avg}} \\
\midrule

\pairband{Pair 1\quad SmolLM3-3B-Base $\leftrightarrow$ Qwen3-1.7B-Base}
\modelband{SmolLM3-3B-Base}
\quad GRPO ($n{=}8$)            & 72.08\stdv{0.52} & 8.58\stdv{1.13} & 8.50\stdv{0.96} & 46.62\stdv{1.49} & 27.22\stdv{0.63} & 32.60\stdv{0.49} & \refrow \\
\oursrow \quad \method{}            & 76.86\stdv{0.32} & 14.42\stdv{1.49} & 14.50\stdv{0.99} & 50.87\stdv{1.60} & 28.68\stdv{0.57} & 37.06\stdv{0.67} & \up{4.46} \\
\quad \method{} w/ stored traj.    & 74.53\stdv{0.41} & 12.00\stdv{1.62} & 12.50\stdv{0.29} & 49.94\stdv{2.36} & 27.44\stdv{0.69} & 35.28\stdv{0.33} & \up{2.68} \\
\addlinespace[3pt]
\modelband{Qwen3-1.7B-Base}
\quad GRPO ($n{=}8$)            & 70.69\stdv{0.44} & 9.17\stdv{1.28} & 4.75\stdv{0.91} & 43.50\stdv{1.57} & 27.89\stdv{0.29} & 31.20\stdv{0.27} & \refrow \\
\oursrow \quad \method{}            & 72.22\stdv{0.25} & 12.50\stdv{0.42} & 7.67\stdv{1.34} & 45.56\stdv{1.09} & 29.23\stdv{0.32} & 33.44\stdv{0.23} & \up{2.24} \\
\quad \method{} w/ stored traj.    & 72.81\stdv{0.63} & 10.92\stdv{0.90} & 5.83\stdv{1.21} & 46.00\stdv{1.42} & 29.49\stdv{0.70} & 33.01\stdv{0.53} & \up{1.81} \\
\midrule

\pairband{Pair 2\quad OctoThinker-3B-Hybrid-Base $\leftrightarrow$ Qwen3-1.7B-Base}
\modelband{OctoThinker-3B-Hybrid-Base}
\quad GRPO ($n{=}8$)            & 56.32\stdv{0.60} & 2.50\stdv{0.93} & 1.08\stdv{0.86} & 27.94\stdv{1.75} & 18.06\stdv{0.81} & 21.18\stdv{0.40} & \refrow \\
\oursrow \quad \method{}            & 58.01\stdv{0.36} & 3.42\stdv{0.80} & 1.25\stdv{0.78} & 34.75\stdv{2.98} & 19.90\stdv{0.71} & 23.47\stdv{0.68} & \up{2.29} \\
\quad \method{} w/ stored traj.    & 57.91\stdv{0.83} & 4.00\stdv{1.76} & 2.42\stdv{0.85} & 31.06\stdv{2.28} & 20.03\stdv{0.51} & 23.08\stdv{0.35} & \up{1.90} \\
\addlinespace[3pt]
\modelband{Qwen3-1.7B-Base}
\quad GRPO ($n{=}8$)            & 70.69\stdv{0.44} & 9.17\stdv{1.28} & 4.75\stdv{0.91} & 43.50\stdv{1.57} & 27.89\stdv{0.29} & 31.20\stdv{0.27} & \refrow \\
\oursrow \quad \method{}            & 71.75\stdv{0.36} & 11.33\stdv{0.99} & 7.25\stdv{1.20} & 44.56\stdv{3.88} & 29.02\stdv{0.77} & 32.78\stdv{0.49} & \up{1.58} \\
\quad \method{} w/ stored traj.    & 71.86\stdv{0.24} & 11.08\stdv{1.46} & 7.08\stdv{0.98} & 44.06\stdv{2.17} & 29.03\stdv{0.33} & 32.63\stdv{0.63} & \up{1.43} \\
\midrule

\pairband{Pair 3\quad SmolLM3-3B-Base $\leftrightarrow$ OctoThinker-3B-Hybrid-Base}
\modelband{SmolLM3-3B-Base}
\quad GRPO ($n{=}8$)            & 72.08\stdv{0.52} & 8.58\stdv{1.13} & 8.50\stdv{0.96} & 46.62\stdv{1.49} & 27.22\stdv{0.63} & 32.60\stdv{0.49} & \refrow \\
\oursrow \quad \method{}            & 73.28\stdv{0.14} & 10.42\stdv{1.02} & 10.83\stdv{0.78} & 46.44\stdv{1.49} & 25.35\stdv{0.24} & 33.26\stdv{0.32} & \up{0.66} \\
\quad \method{} w/ stored traj.    & 73.83\stdv{0.36} & 10.42\stdv{1.14} & 12.67\stdv{1.05} & 48.56\stdv{2.51} & 27.83\stdv{0.76} & 34.66\stdv{0.62} & \up{2.06} \\
\addlinespace[3pt]
\modelband{OctoThinker-3B-Hybrid-Base}
\quad GRPO ($n{=}8$)            & 56.32\stdv{0.60} & 2.50\stdv{0.93} & 1.08\stdv{0.86} & 27.94\stdv{1.75} & 18.06\stdv{0.81} & 21.18\stdv{0.40} & \refrow \\
\oursrow \quad \method{}            & 57.59\stdv{0.50} & 4.25\stdv{0.75} & 2.25\stdv{0.70} & 30.19\stdv{1.30} & 18.74\stdv{0.76} & 22.60\stdv{0.22} & \up{1.42} \\
\quad \method{} w/ stored traj.    & 55.21\stdv{0.56} & 3.50\stdv{1.05} & 2.25\stdv{0.76} & 30.63\stdv{2.80} & 18.31\stdv{0.91} & 21.98\stdv{0.33} & \up{0.80} \\
\bottomrule
\end{tabular}
}
\end{table}

\clearpage
\section{Full Results for Ablations and Alternative Designs}
\label{app:ablation-full}

\Cref{tab:ablation-full} reports per-benchmark results for \Cref{tab:ablation-comp,tab:ablation-design}, and the compatibility-threshold sweep.
\Cref{tab:which-how-full} reports per-benchmark results for combinations of prior methods with \method{}'s prompt selection or peer update.
Implementation details are provided in Appendix~\ref{app:alt-designs}.

All listed variants lower the aggregate score relative to full \method{} for both models, although individual benchmark scores can improve.
Removing compatibility weighting causes the largest aggregate degradation for SmolLM3, whereas removing only the floor causes the largest degradation for Qwen3 among the variants in \Cref{tab:ablation-full}.
Count-matched random selection and alternative peer-minibatch orderings also reduce aggregate performance.
Among the tested thresholds, $\delta=0.8$ achieves the highest aggregate score for both receivers.

\begin{table}[H]
\centering
\small
\setlength{\tabcolsep}{5pt}
\caption{\textbf{Full ablation of \method{} on Pair~1.}
Per-benchmark scores behind \Cref{tab:ablation}.
Each variant changes a single component of the full method while keeping the compute budget fixed at $n{=}8$ rollouts per model.
\emph{Count-matched random} variants match the prompt-selection or response-admission count obtained by applying \method{} to the current rollout batch, separately in each direction, but select uniformly at random.
Random admission replaces the compatibility floor while retaining weights $\min\{s,1\}$.
$\Delta$ reports the change in average score relative to full \method{} within each model block.}
\label{tab:ablation-full}
\resizebox{\linewidth}{!}{
\begin{tabular}{l c c c c c c c}
\toprule
\textbf{Variant} & {MATH500} & {AIME2024} & {AIME2025} & {AMC23} & {Minerva} & {\textbf{Avg.}} & {$\Delta$\textsubscript{Avg}} \\
\midrule
\modelband{SmolLM3-3B-Base}
\oursrow \quad \method{} (full, $\delta{=}0.8$)     & 76.86\stdv{0.32} & 14.42\stdv{1.49} & 14.50\stdv{0.99} & 50.87\stdv{1.60} & 28.68\stdv{0.57} & 37.06\stdv{0.67} & \refrow \\
\addlinespace[2pt]
\multicolumn{8}{l}{\emph{Removing one component}} \\
\quad w/o compatibility gate                     & 68.66\stdv{0.40} & 5.42\stdv{1.47}  & 4.92\stdv{1.12}  & 39.19\stdv{1.66} & 25.33\stdv{0.57} & 28.70\stdv{0.78} & \dn{8.36} \\
\quad w/o balanced exchange                      & 75.05\stdv{0.25} & 10.08\stdv{1.68} & 12.50\stdv{0.78} & 49.44\stdv{2.20} & 28.24\stdv{0.48} & 35.06\stdv{0.27} & \dn{2.00} \\
\quad w/o token-level ratio                      & 75.83\stdv{0.58} & 10.50\stdv{1.04} & 12.50\stdv{1.56} & 48.94\stdv{1.80} & 28.26\stdv{0.49} & 35.21\stdv{0.65} & \dn{1.85} \\
\addlinespace[2pt]
\multicolumn{8}{l}{\emph{Count-matched random selection}} \\                                  
\quad random prompts                             & 75.81\stdv{0.31} & 10.08\stdv{1.12} & 11.75\stdv{0.68} & 47.94\stdv{1.69} & 27.70\stdv{0.71} & 34.66\stdv{0.47} & \dn{2.40} \\
\quad random admission                           & 76.00\stdv{0.45} & 11.25\stdv{1.53} & 11.25\stdv{1.67} & 50.25\stdv{3.02} & 27.24\stdv{0.53} & 35.20\stdv{0.85} & \dn{1.86} \\
\addlinespace[2pt]
\multicolumn{8}{l}{\emph{Position of peer minibatches}} \\ 
\quad first                                      & 75.74\stdv{0.32} & 9.83\stdv{1.49}  & 11.25\stdv{0.51} & 49.06\stdv{1.98} & 27.18\stdv{0.53} & 34.61\stdv{0.43} & \dn{2.45} \\
\quad uniform                                    & 72.47\stdv{0.68} & 12.58\stdv{1.30} & 9.92\stdv{0.90} & 45.06\stdv{2.53} & 27.42\stdv{0.63} & 33.49\stdv{0.47} & \dn{3.57} \\
\addlinespace[2pt]
\multicolumn{8}{l}{\emph{Advantage computation}} \\
\quad pooled with receiver group                 & 74.21\stdv{0.44} & 9.58\stdv{1.32} & 9.17\stdv{1.06} & 45.88\stdv{1.39} & 26.53\stdv{0.61} & 33.07\stdv{0.47} & \dn{3.99} \\
\addlinespace[2pt]
\multicolumn{8}{l}{\emph{Varying the compatibility gate threshold}} \\
\quad $\delta{=}0$ (no floor)                    & 75.66\stdv{0.49} & 11.75\stdv{1.23} & 12.17\stdv{1.16} & 49.38\stdv{1.86} & 26.89\stdv{0.66} & 35.17\stdv{0.59} & \dn{1.89} \\
\quad $\delta{=}0.7$                             & 74.16\stdv{0.48} & 11.17\stdv{1.54} & 9.58\stdv{1.06}  & 48.69\stdv{1.16} & 27.26\stdv{0.88} & 34.17\stdv{0.49} & \dn{2.89} \\
\quad $\delta{=}0.9$                             & 77.38\stdv{0.26} & 9.83\stdv{1.43}  & 14.00\stdv{0.48} & 51.75\stdv{1.65} & 27.75\stdv{0.65} & 36.14\stdv{0.18} & \dn{0.92} \\
\addlinespace[2pt]
\quad GRPO ($n{=}8$)                             & 72.08\stdv{0.52} & 8.58\stdv{1.13}  & 8.50\stdv{0.96}  & 46.62\stdv{1.49} & 27.22\stdv{0.63} & 32.60\stdv{0.49} & \dn{4.46} \\
\addlinespace[3pt]
\modelband{Qwen3-1.7B-Base}
\oursrow \quad \method{} (full, $\delta{=}0.8$)     & 72.22\stdv{0.25} & 12.50\stdv{0.42} & 7.67\stdv{1.34}  & 45.56\stdv{1.09} & 29.23\stdv{0.32} & 33.44\stdv{0.23} & \refrow \\
\addlinespace[2pt]
\multicolumn{8}{l}{\emph{Removing one component}} \\
\quad w/o compatibility gate                     & 69.20\stdv{0.25} & 8.00\stdv{0.80}  & 5.75\stdv{1.12}  & 42.81\stdv{1.89} & 28.88\stdv{0.37} & 30.93\stdv{0.52} & \dn{2.51} \\
\quad w/o balanced exchange                      & 71.77\stdv{0.59} & 11.25\stdv{1.14} & 7.67\stdv{0.48}  & 44.94\stdv{2.01} & 29.17\stdv{0.33} & 32.96\stdv{0.56} & \dn{0.48} \\
\quad w/o token-level ratio                      & 72.18\stdv{0.46} & 11.67\stdv{1.69} & 6.50\stdv{1.09}  & 44.81\stdv{1.44} & 29.77\stdv{0.33} & 32.99\stdv{0.33} & \dn{0.45} \\
\addlinespace[2pt]
\multicolumn{8}{l}{\emph{Count-matched random selection}} \\
\quad random prompts                             & 71.51\stdv{0.29} & 9.50\stdv{1.39} & 6.50\stdv{1.09} & 43.44\stdv{1.29} & 28.35\stdv{0.56} & 31.86\stdv{0.48} & \dn{1.58} \\
\quad random admission                           & 71.29\stdv{0.29} & 11.08\stdv{1.09} & 6.50\stdv{1.52} & 44.56\stdv{1.03} & 28.29\stdv{0.39} & 32.35\stdv{0.32} & \dn{1.09} \\
\addlinespace[2pt]
\multicolumn{8}{l}{\emph{Position of peer minibatches}} \\
\quad first                                      & 71.29\stdv{0.53} & 9.75\stdv{0.56}  & 5.42\stdv{0.88}  & 43.69\stdv{2.33} & 28.48\stdv{0.36} & 31.73\stdv{0.58} & \dn{1.71} \\
\quad uniform                                    & 71.25\stdv{0.28} & 11.08\stdv{1.09} & 6.58\stdv{0.75} & 44.31\stdv{1.88} & 28.80\stdv{0.43} & 32.41\stdv{0.55} & \dn{1.03} \\
\addlinespace[2pt]
\multicolumn{8}{l}{\emph{Advantage computation}} \\
\quad pooled with receiver group                 & 69.77\stdv{0.70} & 8.92\stdv{1.00} & 5.50\stdv{0.90} & 41.88\stdv{1.45} & 27.87\stdv{0.58} & 30.78\stdv{0.48} & \dn{2.66} \\
\addlinespace[2pt]
\multicolumn{8}{l}{\emph{Varying the compatibility gate threshold}} \\
\quad $\delta{=}0$ (no floor)                    & 66.97\stdv{0.60} & 8.50\stdv{1.05}  & 4.58\stdv{1.18}  & 41.69\stdv{1.87} & 27.66\stdv{0.73} & 29.88\stdv{0.62} & \dn{3.56} \\
\quad $\delta{=}0.7$                             & 71.94\stdv{0.21} & 10.83\stdv{0.78} & 7.00\stdv{1.30}  & 46.31\stdv{1.87} & 28.41\stdv{0.79} & 32.90\stdv{0.50} & \dn{0.54} \\
\quad $\delta{=}0.9$                             & 71.33\stdv{0.66} & 9.58\stdv{1.95}  & 6.08\stdv{1.83}  & 43.38\stdv{1.91} & 29.14\stdv{0.50} & 31.90\stdv{0.63} & \dn{1.54} \\
\addlinespace[2pt]
\quad GRPO ($n{=}8$)                             & 70.69\stdv{0.44} & 9.17\stdv{1.28}  & 4.75\stdv{0.91}  & 43.50\stdv{1.57} & 27.89\stdv{0.29} & 31.20\stdv{0.27} & \dn{2.24} \\
\bottomrule
\end{tabular}}
\end{table}

\begin{table}[H]
\centering
\small
\setlength{\tabcolsep}{5pt}
\caption{\textbf{Full results for prior methods with one side replaced, on Pair~1.}
Per-benchmark results for combinations of prior methods with \method{}'s prompt selection or peer update.
HACPO and SGT rows are from \Cref{tab:main-results}; ``SGT $+$ \method{} how'' is identical to ``w/o balanced exchange'' in \Cref{tab:ablation-full}.
$\Delta$ reports the change in average score relative to full \method{} within each model block.}
\label{tab:which-how-full}
\resizebox{\linewidth}{!}{
\begin{tabular}{l c c c c c c c}
\toprule
\textbf{Method} & {MATH500} & {AIME2024} & {AIME2025} & {AMC23} & {Minerva} & {\textbf{Avg.}} & {$\Delta$\textsubscript{Avg}} \\
\midrule
\modelband{SmolLM3-3B-Base}
\oursrow \quad \method{}                  & 76.86\stdv{0.32} & 14.42\stdv{1.49} & 14.50\stdv{0.99} & 50.87\stdv{1.60} & 28.68\stdv{0.57} & 37.06\stdv{0.67} & \refrow \\
\addlinespace[2pt]
\quad HACPO                               & 69.58\stdv{0.53} & 7.08\stdv{1.79}  & 10.83\stdv{1.69} & 42.88\stdv{1.61} & 27.17\stdv{0.61} & 31.51\stdv{0.43} & \dn{5.55} \\
\quad HACPO $+$ \method{} which           & 67.73\stdv{0.43} & 6.42\stdv{0.37} & 6.42\stdv{1.16} & 40.06\stdv{1.64} & 24.23\stdv{0.58} & 28.97\stdv{0.52} & \dn{8.09} \\
\addlinespace[2pt]
\quad SGT                                 & 75.53\stdv{0.51} & 10.67\stdv{1.52} & 11.50\stdv{0.37} & 49.06\stdv{1.51} & 27.63\stdv{0.68} & 34.88\stdv{0.21} & \dn{2.18} \\
\quad SGT $+$ \method{} how               & 75.05\stdv{0.25} & 10.08\stdv{1.68} & 12.50\stdv{0.78} & 49.44\stdv{2.20} & 28.24\stdv{0.48} & 35.06\stdv{0.27} & \dn{2.00} \\
\quad SGT $+$ \method{} which               & 75.57\stdv{0.39} & 9.08\stdv{1.94} & 12.08\stdv{1.59} & 46.63\stdv{1.49} & 27.02\stdv{0.66} & 34.08\stdv{0.15} & \dn{2.98} \\
\addlinespace[2pt]
\quad \method{} $+$ LUFFY update   & 73.62\stdv{0.45} & 9.50\stdv{0.90} & 7.75\stdv{1.40} & 45.19\stdv{2.28} & 26.50\stdv{0.39} & 32.51\stdv{0.89} & \dn{4.55} \\
\addlinespace[2pt]
\quad GRPO ($n{=}8$)                      & 72.08\stdv{0.52} & 8.58\stdv{1.13}  & 8.50\stdv{0.96}  & 46.62\stdv{1.49} & 27.22\stdv{0.63} & 32.60\stdv{0.49} & \dn{4.46} \\
\addlinespace[3pt]
\modelband{Qwen3-1.7B-Base}
\oursrow \quad \method{}                  & 72.22\stdv{0.25} & 12.50\stdv{0.42} & 7.67\stdv{1.34}  & 45.56\stdv{1.09} & 29.23\stdv{0.32} & 33.44\stdv{0.23} & \refrow \\
\addlinespace[2pt]
\quad HACPO                               & 63.74\stdv{0.60} & 7.17\stdv{0.90}  & 4.17\stdv{1.18}  & 36.06\stdv{1.20} & 25.04\stdv{0.44} & 27.24\stdv{0.30} & \dn{6.20} \\
\quad HACPO $+$ \method{} which           & 65.59\stdv{0.38} & 6.25\stdv{1.28} & 4.92\stdv{1.60} & 35.75\stdv{2.06} & 26.45\stdv{0.45} & 27.79\stdv{0.38} & \dn{5.65} \\
\addlinespace[2pt]
\quad SGT                                 & 70.97\stdv{0.77} & 10.17\stdv{1.34} & 5.92\stdv{0.85}  & 40.75\stdv{0.87} & 28.39\stdv{0.43} & 31.24\stdv{0.56} & \dn{2.20} \\
\quad SGT $+$ \method{} how               & 71.77\stdv{0.59} & 11.25\stdv{1.14} & 7.67\stdv{0.48}  & 44.94\stdv{2.01} & 29.17\stdv{0.33} & 32.96\stdv{0.56} & \dn{0.48} \\
\quad SGT $+$ \method{} which               & 71.00\stdv{0.50} & 10.58\stdv{1.37} & 5.83\stdv{0.78}  & 43.12\stdv{1.50} & 28.90\stdv{1.36} & 31.89\stdv{0.37} & \dn{1.55} \\
\addlinespace[2pt]
\quad \method{} $+$ LUFFY update     & 71.16\stdv{0.37} & 9.42\stdv{1.52} & 5.08\stdv{0.95} & 43.44\stdv{1.47} & 27.94\stdv{0.66} & 31.41\stdv{0.50} & \dn{2.03} \\
\addlinespace[2pt]
\quad GRPO ($n{=}8$)                      & 70.69\stdv{0.44} & 9.17\stdv{1.28}  & 4.75\stdv{0.91}  & 43.50\stdv{1.57} & 27.89\stdv{0.29} & 31.20\stdv{0.27} & \dn{2.24} \\
\bottomrule
\end{tabular}}
\end{table}

\clearpage
\section{Implementation of Alternative Designs}
\label{app:alt-designs}
 
All variants in \Cref{tab:ablation,tab:which-how-full} use Pair~1 and the configuration of \Cref{tab:training-hparams}, and differ from full \method{} or from the corresponding baseline only as described below.
 
\paragraph{HACPO $+$ \method{} which.}
We keep the HACPO configuration of Appendix~\ref{app:hparams} but apply its peer-rollout loss only on prompts where the receiver fails on all $n$ rollouts ($k_B(q){=}0$) and the peer succeeds at least once ($1 \le k_A(q) < n$), subject to the balanced exchange of \Cref{sec:which}. Peer responses on all other prompts receive zero advantage and contribute no gradient.
 
\paragraph{SGT $+$ \method{} how.}
This variant retains SGT's unbalanced prompt-selection rule: the receiver has no successful rollout and the peer has at least one.
For the selected prompts, it uses \method{}'s peer-data learning rule, replacing the receiver group with the full peer group and applying source-computed advantages, compatibility weighting, token-level clipping, and peer-last updates.
Thus, it preserves SGT's prompt eligibility rather than its single-success response selection.
Peer groups with $k_A(q)=n$ have zero source-computed advantages and provide no reward-based policy gradient.
The reported variant corresponds to \method{} without balanced exchange.

\paragraph{SGT $+$ \method{} which.}
We retain SGT's auxiliary supervised objective and loss coefficient, but restrict transfer to prompts selected by \method{}'s complementary and balanced selection rule. For each selected prompt, we uniformly sample one successful peer response for the supervised loss.

\paragraph{LUFFY-style peer updates.}
Following LUFFY~\citep{NEURIPS2025_a9d5c33e}, peer tokens are optimized with the regularized importance-sampling objective $f\bigl(\pi_\theta(o_t \mid q, o_{<t})\bigr)\hat{a}$ with the shaping function $f(x) = x/(x+\gamma)$ and $\gamma = 0.1$, in place of the compatibility gate and the token-level clipped ratio. As in LUFFY, the behavior probability of peer tokens is set to one and no clipping is applied to peer tokens, while receiver tokens use the standard clipped surrogate. Selection, balanced exchange, source-computed advantages, and peer-last ordering follow \method{}.

\paragraph{Count-matched random selection.}
At each step, we apply \method{}'s selection rules to the current rollout batch to determine the selection count separately for each transfer direction.
For \emph{random prompts}, we sample the same number of prompts uniformly from the prompt batch, irrespective of either model's rewards, and transfer their peer groups.
The standard compatibility weighting is then applied to these responses.

For \emph{random admission}, we retain \method{}'s prompt selection and count the peer responses satisfying $s(o\mid q)>\delta$.
We then sample exactly this many responses uniformly from the transferred peer groups.
If $\mathcal{A}_{\mathrm{rand}}$ denotes the randomly selected responses, their weights are
\begin{equation}
w_{\mathrm{random}}(o)=
\mathbf{1}[o\in\mathcal{A}_{\mathrm{rand}}]
\min\{s(o\mid q),1\}.
\end{equation}
The compatibility floor is used only to determine the admission count; it is not applied to the randomly selected responses.
Thus, a selected response with $s(o\mid q)\leq\delta$ retains its bounded compatibility weight.
All other settings follow \method{}.

\paragraph{Peer minibatch position.}
\method{} places minibatches containing peer groups last. \emph{First} places them before the receiver's own minibatches, and \emph{uniform} shuffles the minibatch order uniformly at random at every step.
 
\paragraph{Pooled groups.}
Instead of replacing the receiver's all-fail group, this variant retains both the receiver and peer responses and computes advantages over their concatenation:
\begin{equation}
\mathcal{G}_{\mathrm{pool}}(q)=
\mathcal{G}_B(q)\mathbin{\Vert}\mathcal{G}_A(q),
\qquad
\hat{a}_i=
\frac{r_i-\operatorname{mean}(\mathcal{G}_{\mathrm{pool}})}
{\operatorname{std}(\mathcal{G}_{\mathrm{pool}})+\epsilon_{\mathrm{adv}}}.
\end{equation}
Here, $\Vert$ denotes concatenation and $\epsilon_{\mathrm{adv}}=10^{-6}$.
Since $k_B(q)=0$, the pooled reward mean is $k_A(q)/(2n)$, and the receiver's failed responses receive negative advantages.
Peer responses retain compatibility weights $w(o)$, while receiver responses have unit weight.
All retained response tokens are included in the token-mean loss denominator.
This variant changes the advantage normalization, the responses included in optimization, and the loss normalization; it therefore evaluates pooled group construction as a whole.

\paragraph{Success-only transfer.}
This variant uses the same prompt selection as full \method{} and computes advantages from the original peer group before filtering.
Only successful peer responses contribute to the peer policy-gradient term, with their source-computed advantages retained. Unsuccessful peer responses are excluded from both the policy-gradient numerator and the token-mean loss denominator.

\paragraph{Compatibility gating and floor.}
The \emph{w/o compatibility gate} variant sets $w(o)=1$ for every transferred peer response, removing both the floor and compatibility-dependent weighting.
The \emph{no-floor} variant removes only the admission threshold and uses $w(o)=\min\{s(o\mid q),1\}$.
It therefore retains bounded compatibility weighting while admitting every transferred response.

\section{Clipping Dynamics under Peer Minibatch Position}
\label{app:clip-dynamics}

We examine whether peer-minibatch ordering changes the activation of token-level clipping on Pair~1. Over training steps 1--48, we measure the fraction of tokens for which the clipped surrogate is strictly smaller than the unclipped surrogate: $\hat a_j>0$ with $\rho_{j,t}>1+\varepsilon_{\mathrm{high}}$, or $\hat a_j<0$ with $\rho_{j,t}<1-\varepsilon_{\mathrm{low}}$. We report these fractions separately for self-generated and grafted responses, using one run per ordering (Table~\ref{tab:clipfrac_position}).

\begin{table}[H]
\centering
\small
\setlength{\tabcolsep}{6pt}
\caption{Token-level clipping under different peer-minibatch orderings on Pair~1, measured over training steps 1--48.}
\label{tab:clipfrac_position}
\vspace{2pt}
\begin{tabular}{llccc}
\toprule
& & \multicolumn{3}{c}{Peer minibatch position} \\
\cmidrule(lr){3-5}
Clipped-token fraction (\%) & Update & First & Uniform & Last (ours) \\
\midrule
\multirow{2}{*}{Self-generated}
 & SmolLM3-3B   & 0.046 & 0.043 & 0.045 \\
 & Qwen3-1.7B   & 0.068 & 0.068 & 0.072 \\
\addlinespace
\multirow{2}{*}{Grafted (peer)}
 & SmolLM3-3B\,$\leftarrow$\,Qwen3-1.7B & 0.004 & 0.130 & 0.277 \\
 & Qwen3-1.7B\,$\leftarrow$\,SmolLM3-3B & 0.002 & 0.248 & 0.440 \\
\midrule
\multicolumn{2}{l}{Steps with zero clipping on peer tokens (\%)}
 & 97.6 & 13.0 & 0.0 \\
\bottomrule
\end{tabular}
\end{table}

Peer-last ordering yields the highest clipped-token fraction on grafted responses (0.28--0.44\%), compared with 0.13--0.25\% under uniform ordering and less than 0.01\% under peer-first. This pattern is consistent with earlier receiver updates moving peer-token ratios away from one before peer optimization.

\clearpage

\section{Qualitative Analysis}
\label{app:graft-qualitative}

\begin{figure}[h]
    \centering
    \includegraphics[width=\linewidth]{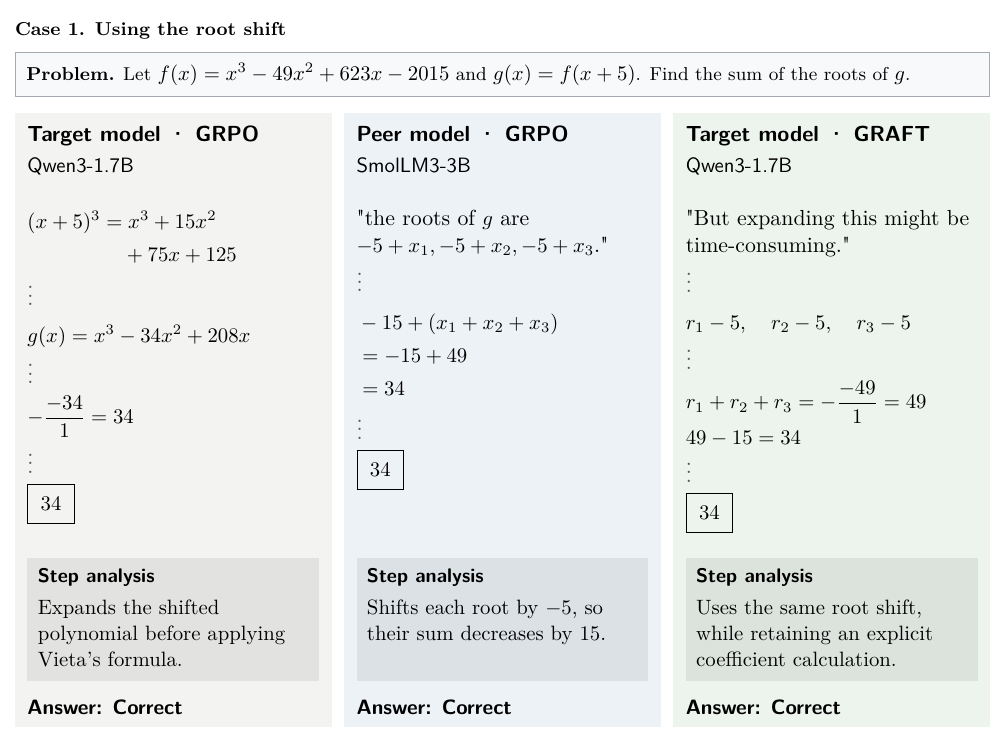}
    \caption{\textbf{Root shift in Qwen3-1.7B (MATH500).}
    Target GRPO expands $g(x)=f(x+5)$ and uses the expanded polynomial to compute the root sum with Vieta's formula. The peer instead shifts each root of $f$ by $-5$, so their sum decreases by $15$. GRAFT uses the same root shift as the peer, computes the original root sum with Vieta's formula, and obtains $49-15=34$.}
    \label{fig:graft-app-q1}
\end{figure}

\begin{figure}[h]
    \centering
    \includegraphics[width=\linewidth]{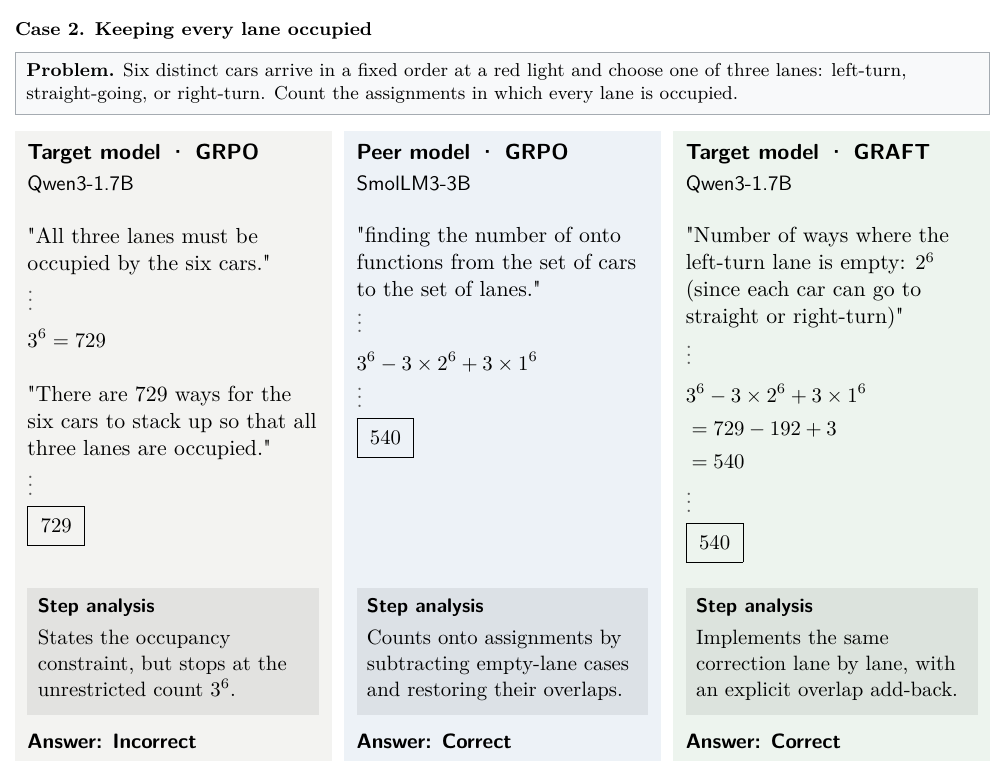}
    \caption{\textbf{Enforcing the occupancy constraint in Qwen3-1.7B (MATH500).}
    Target GRPO counts all $3^6=729$ lane assignments without enforcing that every lane is occupied. The peer treats an assignment as an onto mapping from the six cars to the three lanes and applies inclusion--exclusion. GRAFT applies the same inclusion--exclusion correction through empty-lane events: it subtracts $3\times2^6$ assignments, adds back the three assignments in which only one lane is occupied, and obtains $729-192+3=540$.}
    \label{fig:graft-app-q2}
\end{figure}

\begin{figure}[h]
    \centering
    \includegraphics[width=\linewidth]{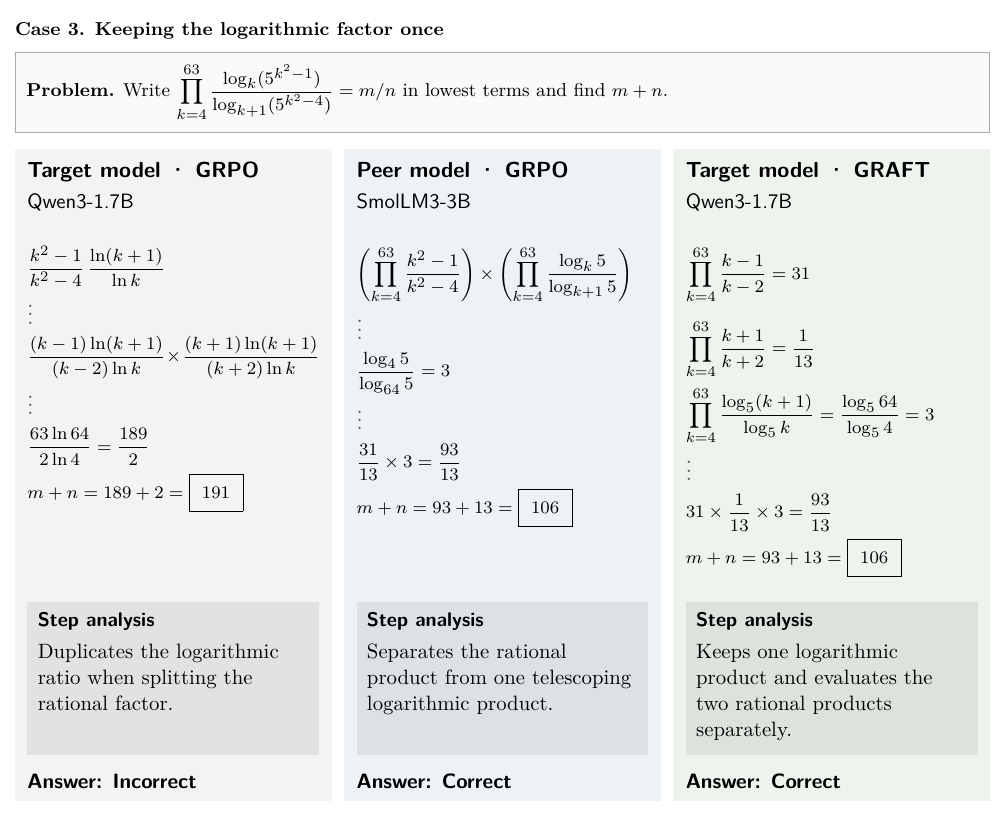}
    \caption{\textbf{Separating the rational and logarithmic factors in Qwen3-1.7B (AIME2025).}
    Target GRPO separates the rational and logarithmic factors, then duplicates the logarithmic ratio when factoring the rational term. The peer separates the rational product from a single logarithmic product that telescopes to $3$. GRAFT uses the same separation with base-$5$ logarithms, evaluates the two rational products as $31$ and $1/13$, and obtains $31\times(1/13)\times3=93/13$, so $m+n=106$.}
    \label{fig:graft-app-q3}
\end{figure}

\begin{figure}[h]
    \centering
    \includegraphics[width=\linewidth]{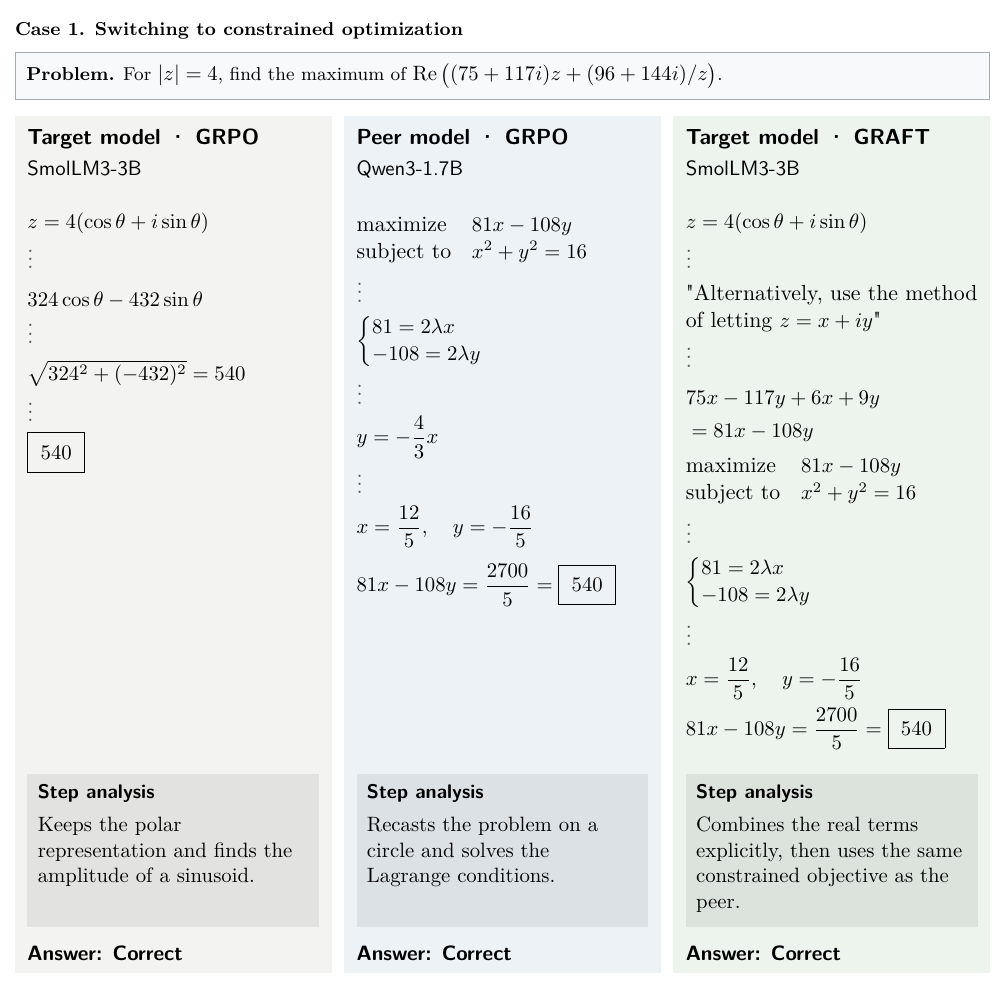}
    \caption{\textbf{Cartesian optimization in SmolLM3-3B (AIME2024).}
    Target GRPO uses the polar parameterization to reduce the objective to $324\cos\theta-432\sin\theta$, whose maximum is $540$. The peer writes $z=x+iy$, rewrites the objective as $81x-108y$ under $x^2+y^2=16$, and solves the constrained problem with Lagrange multipliers. GRAFT starts with the polar parameterization, then switches to $z=x+iy$, derives the same constrained objective as the peer, and obtains $x=12/5$, $y=-16/5$, and the maximum $540$ from the Lagrange equations.}
    \label{fig:graft-app-s1}
\end{figure}

\begin{figure}[h]
    \centering
    \includegraphics[width=\linewidth]{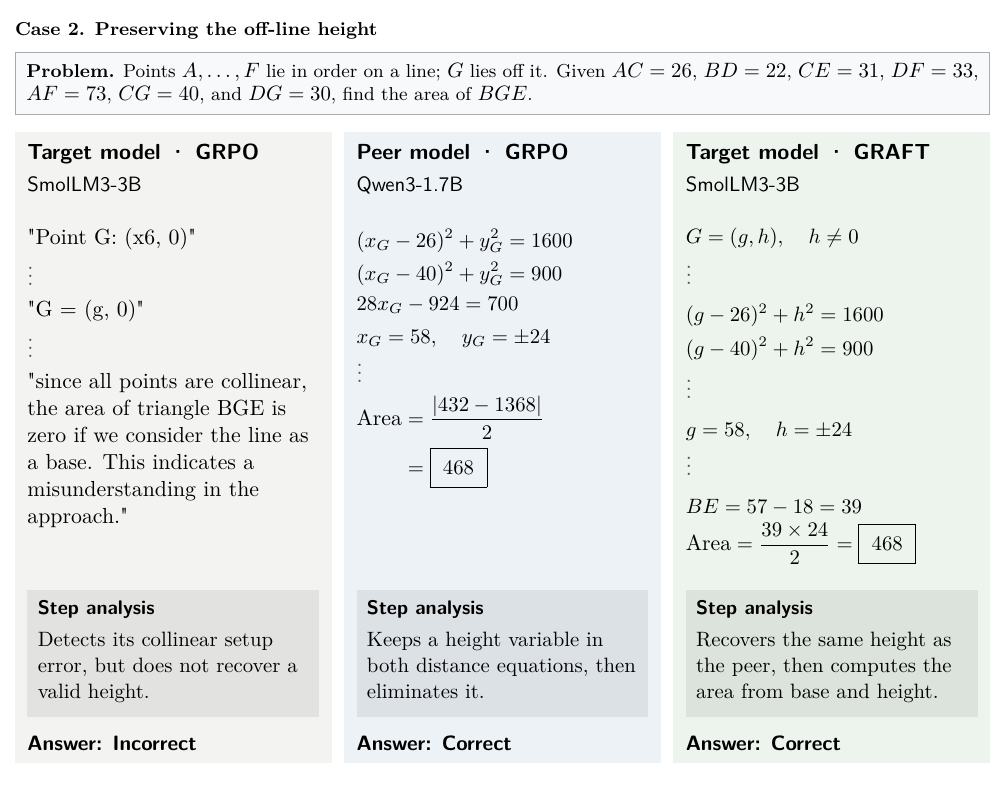}
    \caption{\textbf{Recovering the height in SmolLM3-3B (AIME2025).}
    Target GRPO places $G$ on the line containing $A,\ldots,F$, recognizes the resulting collinearity as an error, but does not recover a nonzero height. The peer keeps a vertical coordinate for $G$ in the distance constraints $CG=40$ and $DG=30$, obtains a height of $24$, and computes the area as $468$ with the shoelace formula. GRAFT places $G$ off the line, uses the two distance constraints to recover a height of $24$, and computes the area from $BE=39$ as $\frac{39 \times 24}{2}=468$.}
    \label{fig:graft-app-s2}
\end{figure}

\begin{figure}[h]
    \centering
    \includegraphics[width=\linewidth]{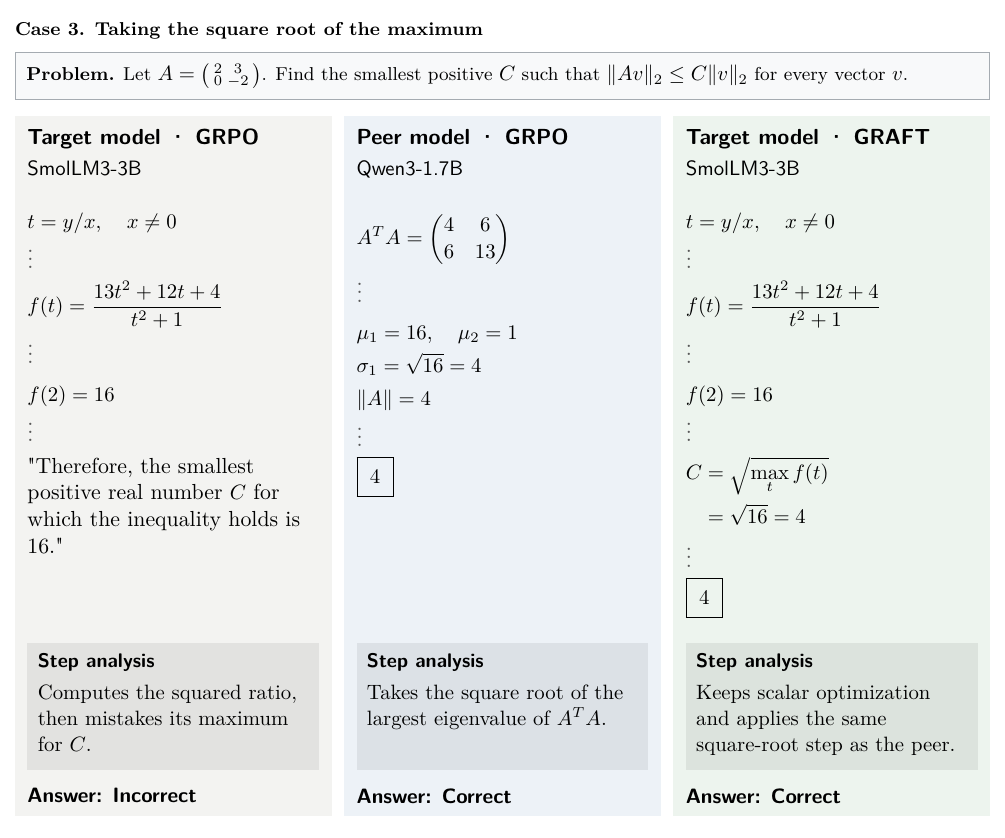}
    \caption{\textbf{Recovering the norm from the squared ratio in SmolLM3-3B (MATH500).}
    Target GRPO maximizes the squared norm ratio $f(t)$ with $t=y/x$, obtains a maximum of $16$, and sets $C=16$ without taking the square root. The peer computes the largest eigenvalue of $A^\top A$ as $16$ and takes its square root to obtain the operator norm $C=4$. GRAFT keeps the scalar optimization used by target GRPO and also takes the square root of the resulting maximum to obtain $C=\sqrt{16}=4$.}
    \label{fig:graft-app-s3}
\end{figure}

We present three examples for each target model in Pair~1, for a total of six examples. Figures~\ref{fig:graft-app-q1}--\ref{fig:graft-app-q3} show Qwen3-1.7B examples and Figures~\ref{fig:graft-app-s1}--\ref{fig:graft-app-s3} show SmolLM3-3B examples. Each figure presents a GRPO response from the target model, a GRPO response from its peer, and a GRAFT response from the target model. We show excerpts of the responses and condense equations for readability. Vertical ellipses mark omitted steps. The shaded \emph{Step analysis} boxes summarize our analysis of the approach taken in each response.
\subsection{Qwen3-1.7B}
\label{app:graft-qualitative-qwen}

\clearpage

\clearpage

\clearpage

\subsection{SmolLM3-3B}
\label{app:graft-qualitative-smol}

\clearpage

\clearpage

\clearpage

\end{document}